\documentclass{article}

\usepackage[square,numbers]{natbib}

\usepackage[final]{neurips_data_2024}

\usepackage[utf8]{inputenc} 
\usepackage[T1]{fontenc}    
\usepackage[pagebackref=true,breaklinks=true,colorlinks,linkcolor=red,citecolor=blue,bookmarks=false]{hyperref}
\usepackage{url}            
\usepackage{booktabs}       
\usepackage{amsfonts}       
\usepackage{nicefrac}       
\usepackage{microtype}      
\usepackage{xcolor}         

\usepackage{wrapfig}
\usepackage{subcaption}
\usepackage{amsmath}
\usepackage{amssymb}
\usepackage{mathtools}
\usepackage{amsthm}
\usepackage{xspace}
\usepackage{bm}
\usepackage{bbm}
\usepackage[capitalize,noabbrev]{cleveref}

\theoremstyle{plain}
\newtheorem{theorem}{Theorem}[section]

\theoremstyle{definition}

\theoremstyle{remark}

\usepackage{xcolor,colortbl}
\usepackage{pifont}         
\usepackage{color, soul}
\definecolor{newred}{HTML}{e31929}
\definecolor{newgreen}{HTML}{018958}
\definecolor{newyellow}{HTML}{f9af2f}

\definecolor{darkergreen}{RGB}{21, 152, 56}
\definecolor{red2}{RGB}{252, 54, 65}

\newcommand{\oursceleb}{CelebA-CF\xspace}
\newcommand{\ourslfw}{LFW-CF\xspace}
\newcommand{\ourscf}{Counterfactual Face\xspace}

\newcommand{\methodname}{CKD\xspace}
\newcommand{\methodnamefull}{Counterfactual Knowledge Distillation\xspace}
\newcommand{\oursmethod}{\methodname}

\definecolor{codegreen}{rgb}{0,0.6,0}
\definecolor{codegray}{rgb}{0.5,0.5,0.5}
\definecolor{codepurple}{rgb}{0.58,0,0.82}
\definecolor{backcolour}{rgb}{0.95,0.95,0.92}

\definecolor{myblue}{RGB}{59,103,188}
\definecolor{mygray}{RGB}{120,120,120}
\definecolor{myyellow}{RGB}{255,184,0}

\definecolor{bestcolor}{RGB}{252, 229, 205}
\newcommand{\best}[1]{\cellcolor{bestcolor}{\textbf{#1}}}
\definecolor{secondcolor}{RGB}{243, 243, 243}
\newcommand{\second}[1]{\cellcolor{secondcolor}{\textbf{#1}}}
\newcommand{\worse}[1]{\textcolor{red}{\textbf{#1}}}

\newcommand{\tteq}{\text{\,\texttt{=}\,}}

\newcommand{\eg}{{\emph{e.g.}}\xspace} 
\newcommand{\ie}{{\emph{i.e.}}\xspace}

\title{Do Counterfactually Fair Image Classifiers Satisfy Group Fairness? -- A Theoretical and Empirical Study}

\author{Sangwon Jung$^{1}$\thanks{Equal contribution.} ~~ Sumin Yu$^{1}$\footnotemark[1] ~~ Sanghyuk Chun$^2$\thanks{Co-corresponding author.}  ~ Taesup Moon$^{1,3}$\footnotemark[2] \vspace{.1in} \\
$^1$ Department of Electrical and Computer Engineering, Seoul National University \\ 
$^2$ NAVER AI Lab \ \ \ \ 
$^3$ ASRI/INMC/IPAI/AIIS, Seoul National University
}

\begin{document}

\maketitle

\begin{abstract}
The notion of algorithmic fairness has been actively explored from various aspects of fairness, such as counterfactual fairness (CF) and group fairness (GF). However, the exact relationship between CF and GF remains to be unclear, especially in image classification tasks; the reason is because we often cannot collect counterfactual samples regarding a sensitive attribute, essential for evaluating CF, from the existing images (\eg, a photo of the same person but with different secondary sex characteristics). In this paper, we construct new image datasets for evaluating CF by using a high-quality image editing method and carefully labeling with human annotators. Our datasets, \oursceleb and \ourslfw, build upon the popular image GF benchmarks; hence, we can evaluate CF and GF simultaneously. We empirically observe that CF does not imply GF in image classification, whereas previous studies on tabular datasets observed the opposite. We theoretically show that it could be due to the existence of a latent attribute $G$ that is correlated with, but not caused by, the sensitive attribute (\eg, secondary sex characteristics are highly correlated with hair length). From this observation, we propose a simple baseline,  Counterfactual Knowledge Distillation (CKD), to mitigate such correlation with the sensitive attributes. 
Extensive experimental results on \oursceleb and \ourslfw demonstrate that CF-achieving models satisfy GF if we successfully reduce the reliance on $G$ (\eg, using CKD). 
\end{abstract}

\section{Introduction}
\label{sec:intro}

As machine learning algorithms are deployed in societal computer vision applications such as facial recognition \cite{wang2019racial} and job interview \cite{nguyen2016hirability}, concerns have grown regarding their potential to discriminate against certain individuals and groups. For instance, a face recognition system might exhibit disparate accuracies across different demographic groups \cite{buolamwini2018gender}, while a job interview algorithm could be biased based on protective attributes even for the same interviewee \cite{objective_biased}. Consequently, \textit{algorithmic fairness} in image classifiers has gained significant attention in academic and industrial research communities.

While conceptually apparent, determining a concrete notion of fairness is challenging, leading to the proposal of several different fairness notions. One prevalent notion is \textit{counterfactual fairness} (CF) \cite{kusner2017counterfactual} which seeks consistent predictions when only a sensitive attribute is intervened. Another important notion is \textit{group fairness} (GF) \cite{zafar2017fairness} that aims to treat different demographic groups equally to prevent one group unfairly disadvantaged compared to another.
Many researchers have focused on developing separate algorithms to achieve each notion, while understanding the exact relationship between CF and GF is yet under-explored; \textit{e.g.,} 
some recent work
\cite{anthis2023causal,rosenblatt2023counterfactual} showed that a model achieving CF can meet several GF notions \textit{only} under specific conditions of Structural Causal Models.

Furthermore, previous studies on  
the relationship between CF and GF have not considered the setting of image classification due to the absence of \textit{evaluation} datasets with counterfactual images, in which only the sensitive attribute is altered from the original images while other attributes not caused by the sensitive attribute remain unchanged --- a data nearly impossible to collect in the real world. 
There have been several works generating counterfactual images using generative models \cite{dash2022evaluating, kim2021counterfactual, louizos2017causal,pfohl2019counterfactual, zhang2022fairness, ramaswamy2021fair,d2024improving}, but they have only focused on utilizing generated counterfactual samples for training rather than evaluation. Moreover, these methods often suffer from low-quality counterfactual images generated based on VAE \cite{kingma2013auto} or GAN \cite{goodfellow2014generative}. One notable exception is \citet{liang2023benchmarking}, which offers an evaluation dataset including counterfactual images. However, their images are all synthetic; thus, it is still insufficient to evaluate CF due to distribution shifts from real-world images.

In this paper, we construct CF benchmarks for image classification tasks using high-performing diffusion model-based generative models. Our datasets build upon popular facial benchmark datasets used for evaluating GF, CelebA and LFW, by altering the sensitive attribute with pre-trained InstructPix2Pix (IP2P) \cite{brooks2023instructpix2pix}. We then carefully curate the edited samples by human annotators and verify the reliability of our datasets as counterfactual samples from additional annotators. Note that our datasets, CelebA-\ourscf (\oursceleb) and LFW-\ourscf (\ourslfw), share the same test samples as the original GF benchmarks, enabling the evaluation of both GF and CF.

Using our datasets, we conduct a primitive study on the relationship between CF and GF in image classification, \eg, test whether CF implies GF for image classifiers. 
To that end, we train CF-aware methods \cite{russell2017worlds, garg2019counterfactual} and evaluate them with our datasets using both CF and GF metrics. From the result, we observe that they achieve CF but fail to satisfy GF, contrary to previous findings that CF can imply GF \cite{anthis2023causal, rosenblatt2023counterfactual}. 
We attribute this failure to Structural Causal Models (SCMs) of image generation. Specifically, for an image SCM, a latent attribute $G$ is more likely to exist, which could be correlated with, but not caused by, the sensitive attributes. For example, in the case where the sensitive attribute is the sex of a person in an image, secondary sex characteristics such as beard and hairline are highly correlated with hair length, but it does not mean that such characteristics cause the length of hair. In this scenario, if a model achieving CF relies on the attribute $G$ (\eg, hair length) on its prediction, it could more severely violate GF in the worst case. Therefore, if we can reduce the dependency on $G$ of a CF-aware model, we may achieve both CF and GF. Empirically, we find that a model trained with vanilla cross-entropy loss is more robust to $G$ than a model trained with a CF-aware method. Motivated by this, we propose a simple baseline, named \methodnamefull (\oursmethod), which distills the robustness to $G$ during the original CF-aware optimization. Finally, our extensive experiments using CelebA-CF and LFW-CF demonstrate that CF-achieving models satisfy GF when reducing the reliance on $G$ (\eg, using \oursmethod).



In summary, our contributions are three-fold. Firstly, we construct two new image classification benchmarks for measuring CF, \oursceleb and \ourslfw.
Secondly, using these datasets, we observe the disparity between CF and GF in image classifiers and provide a theoretical rationale; a 
counterfactually fair classifier may not necessarily achieve GF when an additional latent attribute that is correlated with the sensitive attribute exists.
Finally, we propose a simple baseline, \oursmethod, to reduce the sensitivity to such latent attributes of a model, resulting in achieving CF and GF simultaneously.

\vspace{-.2em}
\section{Constructing high-quality counterfactual images}
\label{sec:ctf_contruction}
\vspace{-.2em}
The degree of counterfactual fairness (CF) can be measured by the prediction consistency between an original sample and its corresponding counterfactual (CTF) sample. For a given sample and a sensitive attribute, a CTF sample is defined as the one of which the sensitive attribute is altered while all the other attributes not caused by the sensitive attribute remain the same. 
However, acquiring a CTF sample for an image is challenging. For example, if the sex of a person in an image is the sensitive attribute, obtaining a CTF sample requires changing the secondary sex characteristics of the person such as beard or hairline, while preserving their identity and the other attributes, which is impossible in practice. 
One possible alternative is to generate a virtual face by altering such secondary sex characteristics of the given identity using a high-quality image editing method.


Several previous approaches \cite{kim2021counterfactual,ramaswamy2021fair,xu2018fairgan,zhang2022fairness,kocaoglu2017causalgan} have attempted to generate CTF images by VAE or GAN-based editing methods. However, they have struggled with low image quality or unintended modifications to non-sensitive attributes, rendering them unreliable for evaluating CF. To address such issues, we employ IP2P \cite{brooks2023instructpix2pix}, an advanced diffusion model-based image editing method.
Notably, IP2P can generate high-quality CTF samples by simply adjusting the text instructions without any model retraining.

\begin{figure}[t!]
\centering
\includegraphics[width=.95\columnwidth]{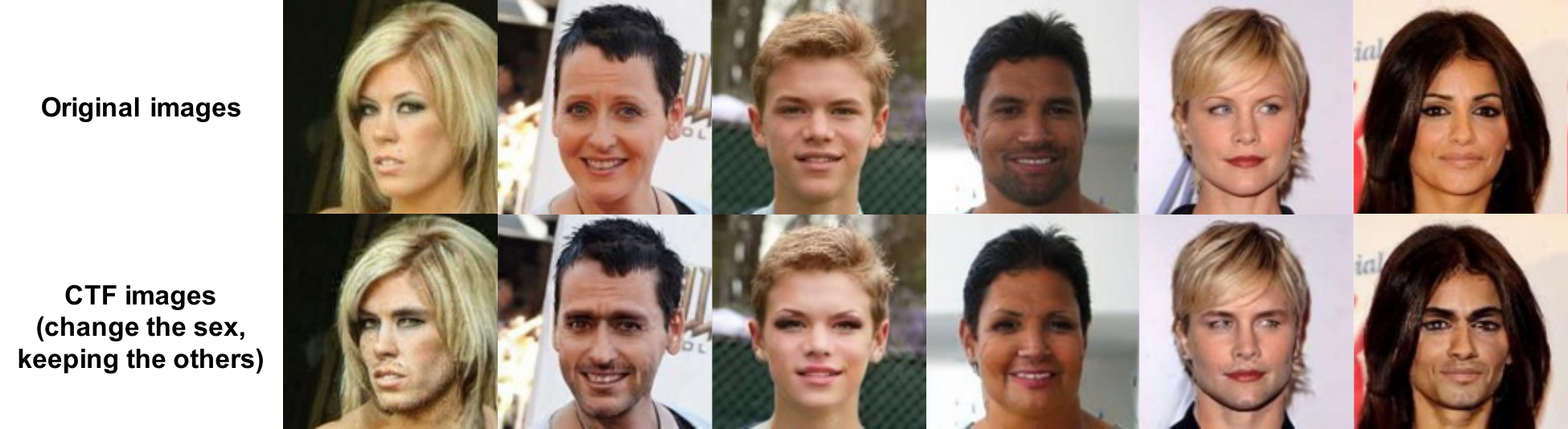}
\caption{\small {\bf \oursceleb examples}. The counterfactual (CTF) images regarding the ``sex attribute'' are shown.}
\label{fig:ctf_example_celeba}
\vspace{-1em}
\end{figure}

As the first step, we edit the test images of two popular facial image datasets, CelebA \cite{celeba} and LFW \cite{Huang2007a}. 
We choose the ``sex'' of a person in an image as the sensitive attribute\footnote{The two datasets use the terms ``gender'' for indicating their sensitive attributes. However, using such terminology can present some ethical concerns because they can suggest meanings linked to social identities. Thus, we have decided to use the term ``sex'' instead, which more accurately refers to biological characteristics.}
and edit the sex-related visual characteristics of facial images using text prompts. We generated 720 CelebA CTF image pairs and 632 LFW CTF image pairs, where the images are selected to be balanced across groups for both target and sensitive labels. Here, we treat ``blond hair'' and ``smiling'' as the target labels for CelebA and LFW, respectively. Namely, for example, the CelebA CTF image pairs have a balanced group of <female, non-blond hair>, \ldots, and so on. \cref{fig:ctf_example_celeba} and \ref{fig:ctf_expample_lfw} show examples of generated CTF images together with the originals. Hyperparameter settings are reported in \cref{appendix:subsec_hyp_ip2p}. Note that while we adopt the ``sex'' attribute, our generation process is attribute-agnostic (\eg, age or skin color can be also used in place of sex) as illustrated in \cref{fig:ctf_example_multi_celeba}.

\noindent\textbf{Image filtering}.
Despite the high quality of IP2P, low-quality CTF images can still be generated. To address this, we employed five human annotators to filter the images, \ie, each image pair was annotated as either  ``correct'' or ``incorrect''. 
To ensure objective and precise annotation criteria, we created guidelines as follows. 
Initially, we compiled a list of 20 masculine and feminine visual features using GPT-4o and with guidance from experts specialized in fairness, selected nine key facial attributes representing sex-related visual characteristics: facial hair, Adam's apple, skin texture, jawline, chin shape, brow ridge, cheekbone prominence, lip fullness, and hairline. These attributes were used to establish the criteria for evaluating correct CTF samples. One notable issue is that most of the feminine-like images in CelebA and LFW datasets include makeups (for instance, many female celebrities in the CelebA dataset appear to be wearing makeup) and the IP2P model is biased towards removing makeup when altering feminine features. To prevent images from being filtered out solely due to changes in makeup, we additionally included makeup in the set of key attributes, even though it is not a sex characteristic. Finally, the guidelines were created based on these ten attributes, providing some criteria for correct CTF samples, such as whether the change of some of the ten attributes was accurate and whether other facial characteristics remained consistent with the original image.  Using these guidelines, we filtered out pairs receiving two or fewer ``correct'' votes, resulting in 230 and 144 images for CelebA and LFW, respectively.
More details about the human annotating interface are in \cref{appendix:human_evaluation}, and additional information on the newly created dataset can be found in \cref{appendix:analysis-dataset}.

\begin{wraptable}{r}{0.4\textwidth}
\centering
\small
\vspace{-2em}
\caption{\small {\bf Human evaluation of the reliability of our datasets}. Accuracies of the correctly altered sensitive attributes and well-preserved non-sensitive attributes are shown. 
}
\label{table:double-check}
\begin{tabular}{lcc}
\toprule
& Sensitive & Non-Sensitive \\
\midrule
\oursceleb & 96.52 & 95.98 \\
\ourslfw & 98.61 & 93.75 \\
\bottomrule
\end{tabular}
\vskip -0.1in
\end{wraptable}
\noindent\textbf{Reliability check}. 
We further verify the quality of our datasets by additional five human annotators, distinct from those participated in the filtering process. Those annotators evaluate only the images that remained after the filtering, based on two criteria: (1) whether the sensitive attribute was correctly changed and (2) whether the other non-sensitive attributes were preserved. The annotators evaluated the images for the sensitive attribute, ``sex'' and three non-sensitive attributes, ``blond hair'', ``gray hair'', and ``smiling''; we chose these three because other attributes can be subjective (\eg, ``big nose'') \cite{wu2023consistency} or had already been filtered (\eg, ``wearing hat'').
Details of the annotating interface provided to the five annotators are in \cref{appendix:human_evaluation}. Based on the majority vote, we compute the percentage of CTF samples which met each of the two criteria, \ie, the accuracies for whether the sensitive and non-sensitive attributes are correctly altered and preserved. \cref{table:double-check} displays the values for \oursceleb and \ourslfw. The non-sensitive accuracy is averaged across three non-sensitive attributes. The results demonstrate that our CTF samples almost meet the two CTF criteria, suggesting that our datasets can be reliably utilized to evaluate CF.

\noindent\textbf{Ethical considerations}.
In our study, we use the term ``sex'', not ``gender'', to represent the sensitive attribute with biological traits, because terms such as ``gender'' might imply associations with social identities, potentially raising some ethical issues. We also specifically choose ten perceived facial attributes as the visual features representing the biological sex in facial images. We believe that these considerations help alleviate various normative harms that arise from dichotomizing gender, which refers to social identity. However, despite our efforts, the sex-related visual characteristics are complex and intertwined, making it challenging to fully represent with a binary label.
Thus, we urge practitioners to use our datasets with these considerations in mind.

\section{Primitive study on the relationship between CF and GF}
\label{sec:pre-exp}
\subsection{Experimental setup}
\label{subsec:setup}
We consider the image classification task where each data sample consists of an input image $X$,
a class attribute $Y \in \mathcal{Y}=\{0, \cdots, |\mathcal{Y}|-1\}$ and a sensitive attribute $A \in \{0, 1\}$, \eg, sex.

\paragraph{Metrics.} 
We measure three metrics for CF, GF, and classification accuracy. Firstly, we describe the metric for CF. A classifier satisfies CF when the predictions for the original sample and its counterfactual (CTF) sample are the same for every sample $x$ and sensitive attribute $a$, \ie, $P(\widehat{Y}\tteq y|X\tteq x, A\tteq a) =P(\widehat{Y}_{A\leftarrow a'}\tteq y|X\tteq x, A\tteq a)$, where $\widehat{Y}_{A\leftarrow a'}$ represents the prediction for a counterfactual sample intervened on $A$ with $a'$ (\eg, changing female to male). We quantify the degree of violence with respect to CF using counterfactual disparity (CD):

{\small
\vspace{-1.5em}
\begin{align}
\text{
\textbf{Counterfactual Fairness:} 
Counterfactual Disparity \textbf{(CD)}}\triangleq\mathbb{E}_{x,a} \big[P\big(\mathbbm{1} \{\widehat{Y}_{A\leftarrow a'}  \neq \widehat{Y}\}|x,a\big)\big]. \label{eq:cd}
\end{align}
\vspace{-1.5em}
}

Secondly, we adopt equalized odds (EO) as our notion for GF. 
If a predictor $\widehat{Y}$ and the sensitive attribute $A$ are conditionally independent given the true class attribute $Y$, the predictor satisfies EO; namely, EO holds when $P(\widehat{Y}\tteq y'|A\tteq 0,Y \tteq y) =P(\widehat{Y}\tteq y'|A\tteq 1,Y\tteq y)$.
From the definition, we can capture the degree of violence with respect to GF with the disparity of EO (DEO):

{\small
\vspace{-1.5em}
\begin{align}
\text{Disparity of EO \textbf{(DEO)}}\triangleq \max_{y,y'\in\mathcal{Y}} \big |P&(\widehat{Y}\tteq y' |A\tteq0, Y\tteq y)\text{\texttt{-}}P(\widehat{Y}\tteq y' |A\tteq1,Y\tteq y)\big|. \label{eq:deo}
\end{align}
\vspace{-1.5em}
}

We note that we empirically compute CD and DEO, defined in \cref{eq:cd} and (\ref{eq:deo}), using our benchmark datasets and the original test datasets of CelebA and LFW, respectively.
Additionally, \citet{pinto2024matrix} propose several other metrics to evaluate CF, and accordingly, we conducted an additional evaluation based on these metrics, with results provided in \cref{appendix:additional-metric-CCM}.

\paragraph{Baseline methods.}
We evaluate a model trained with the vanilla cross-entropy loss (denoted as ``Scratch'') and two CF-aware training methods, Scratch+aug and counterfactual pairing (CP). 
Scratch+aug is a Scratch method using an augmented training dataset with counterfactual samples \cite{garg2019counterfactual}, and CP \cite{russell2017worlds} adopts a regularization term that 
promotes pairs of original and its CTF sample to obtain the same prediction (see \cref{eq:cp} for the formal definition). Note both methods need counterfactual samples for training, and hence, we use the samples generated via IP2P with the same prompts used in \cref{sec:ctf_contruction} without any filtering process to obtain results for them. For a comprehensive study, we additionally evaluate two individual fairness-aware methods, SenSeI \cite{yurochkin2020sensei} and LASSI \cite{peychev2022latent}, of which goals are analogous to CF in aiming to make a model robust to perturbation of the sensitive attribute.
More details are described in \cref{appendix:baselines-details}.

\paragraph{Model selection.} Due to the accuracy-fairness trade-off \cite{dutta2020there}, appropriate model selection is important for fair evaluation. We explore varying hyperparameters and select the best model that shows the lowest CD (\cref{eq:cd}) for the held-out validation set while achieving a lower bound of the accuracy\footnote{Considering the accuracy degradation of fair-training methods, we set the bound as 98\% of the accuracy of Scratch, \ie, if Scratch achieves 95.0\% accuracy, then we only consider models with more than 93.1\% accuracy.}.
\begin{table}[t!]
\centering
\vspace{-.75em}
\caption{\textbf{CF does not always imply GF on image classification}. We report CD (\cref{eq:cd}) and DEO (\cref{eq:deo}) for measuring Counterfactual Fairness (CF) and Group Fairness (GF), respectively. Accuracy and DEO are measured on the original test datasets (CelebA and LFW) and CD is evaluated on the newly constructed datasets, \oursceleb and \ourslfw, described in \cref{sec:ctf_contruction}. If a model shows an inferior metric value than Scratch, the number is highlighted in \worse{red}.}
\label{table:method-comparison}
\setlength{\tabcolsep}{8pt}
\small
\begin{tabular}{lcccccc}
\toprule
& \multicolumn{3}{c}{CelebA (and \oursceleb)} & \multicolumn{3}{c}{LFW (and \ourslfw)} \\
 Method         & Acc $\uparrow$ & CD $\downarrow$ & DEO $\downarrow$ & Acc $\uparrow$ & CD $\downarrow$ & DEO $\downarrow$ \\
\midrule
Scratch        & 95.53   & 10.26  & 47.10    & 90.85   & 18.06 & 7.66  \\ \midrule
Scratch+aug \cite{garg2019counterfactual} & 95.41    & 4.65  & {44.71}   & 90.34   & 12.15  & \worse{7.86} \\
CP \cite{russell2017worlds}    & 94.10    & {2.53}   & \worse{51.01}   & 89.77  & {9.20} & \worse{8.74}    \\
SenSeI \cite{yurochkin2020sensei}     & 95.33   & 8.00   & \worse{52.32}   & 87.75   & 16.09   & \worse{9.23} \\ 
LASSI \cite{peychev2022latent}      & 91.07 &  9.69 & 31.79 & - & - & -  \\\midrule
\end{tabular}
\vspace{-1em}
\end{table}
\subsection{Performance comparison}
\label{subsec:performance-comparison}
\cref{table:method-comparison} shows accuracy, CD, and DEO for Scratch and four baseline methods. Note that we omit the result of LASSI on LFW because the number of samples in LFW is not enough to train the Glow model \cite{kingma2018glow}, which is the main component of LASSI. From the table, CF-aware and individual fairness-aware methods are mostly effective in mitigating CD, when compared to Scratch. However, it does not necessarily lead to improvements in DEO. Especially, while CP significantly improves CD for both datasets, it exacerbates DEO compared to Scratch. Namely, contrary to the previous studies \cite{anthis2023causal, rosenblatt2023counterfactual} showing that CF implies GF on tabular datasets, our observation shows that CF does not always imply GF on image datasets. In the following section, we theoretically investigate why the previous observations may not hold on images.

\section{Theoritical analysis on the relationship between CF and EO for images}
\label{sec:theory}

\begin{wrapfigure}{r}{0.35\textwidth}
    \vspace{-2em}
  \begin{center}
    \includegraphics[width=\linewidth]{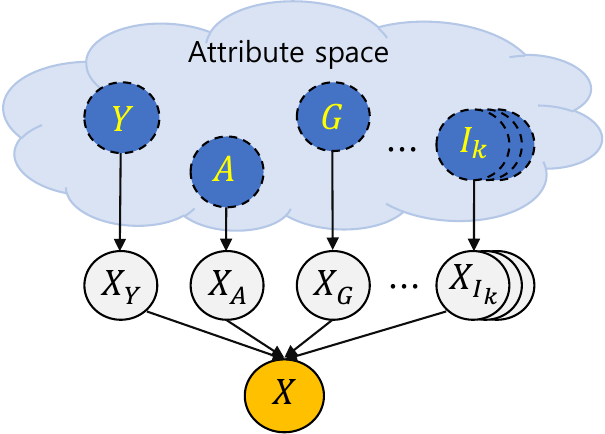}
  \end{center}
  \caption{\small \textbf{Image SCM}. \textcolor{myblue}{Blue}, \textcolor{mygray}{gray}, and \textcolor{myyellow}{yellow} circles represent \textcolor{myblue}{latent attributes}, \textcolor{mygray}{components of an image} and \textcolor{myyellow}{a whole image}, respectively. Directed edges indicate a causal relationship from the source to the target. The blue region indicates that there can be any direction of edges between blue nodes.}
  \label{fig:scm}
  \vspace{-1em}
\end{wrapfigure}
\subsection{Structural Causal Model (SCM) for images}
\label{subsec:scm-vision}
Structural Causal Models (SCMs) are represented as directed acyclic graphs satisfying the conditions specified in \cite{pearl2000models}. In these models, nodes and edges indicate variables and their causal relationships within the data-generating process. As studied in previous works \cite{dash2022evaluating,kocaoglu2017causalgan}, the nodes of an SCM for image can be categorized into three parts. As shown in \cref{fig:scm}, the \textcolor{myblue}{blue}, \textcolor{mygray}{gray} and \textcolor{myyellow}{yellow} nodes indicate \textcolor{myblue}{latent attributes}, \eg, $Y$ or $A$, \textcolor{mygray}{components of the image influenced by these attributes}, \eg, $X_Y$ or $X_A$, and \textcolor{myyellow}{the whole image} $X$, respectively. Taking an SCM for facial images as an example, we can interpret these nodes as follows: latent attributes such as \textcolor{myblue}{hair color or sex}, facial components like the \textcolor{mygray}{hair or an Adam's apple} in a facial image, and \textcolor{myyellow}{an entire face}. Note that the blue region in the figure describes that various causal relationships among latent attributes can exist \footnote{We assume no edge or unblocked path from $A$ to $Y$; otherwise, all counterfactually fair models based on that SCM would produce random predictions with respect to $X_Y$.}. Furthermore, although an image SCM may contain additional latent attributes, we simplify our focus to only include the class and sensitive attribute, $Y$ and $A$, and a third-party attribute, $G$, which may correlate with the sensitive attribute $A$.


\subsection{Theoretical analysis}
\label{subsec:theorem}
According to the Markov assumption of SCM \cite{pearl2000models}, if there are no unblocked paths between two variables in an SCM (\ie, they are d-separated), the variables are statistically independent. Utilizing this property, 
\citet{anthis2023causal} demonstrated that CF implies several GF notions, including Equalized Odds (EO) \cite{hardt2016equality}, under the specific condition on SCMs such as no \textit{backdoor} path from the sensitive attribute $A$ to the image $X$ exists (Theorem 2 of \cite{anthis2023causal}). Moreover, the authors empirically show that these conditions would hold on some tabular datasets.

However, we argue that these conditions would not hold for image datasets due to a fundamental difference in what sensitive attributes represent in an image. Specifically, tabular datasets typically consist of recorded information by subjects, where sensitive attributes such as sex or race usually represent immutable genetic information; hence they are not caused by other attributes and cause all attributes correlated with the sensitive attributes. In contrast, sensitive attributes in image datasets indicate visual characteristics that can change and be influenced by some other attributes, such as the attribute $G$. For example, in a facial image dataset, attributes like hair length or accessories might be highly correlated with, but not caused by, secondary sex characteristics such as beard. Namely, a backdoor path from the sensitive attribute $X$ through the attribute $G$ could exist, thereby breaking the connection between CF and GF discovered in previous studies.


Our theoretical result specifies the relationship between CF and GF (especially for EO) with $G$:
{
\begin{theorem}
\label{thm1}
Assume a latent attribute $G$ in \cref{fig:scm} is a non-descendant variable of $A$ and connected to $A$ through an unblocked path. Then, the following inequality holds for a counterfactually fair classifier $\bm\theta$ and any pairs of $y$ and $y'$:
\begin{align}
\label{eq:theorem}
\big|P(\widehat{Y}=y'|A=0,Y=y)-P(\widehat{Y}=y'|A=1,Y=y)&\big| \nonumber \\ \leq \sum_{X_Y}P(X_Y|Y=y)&\max_{X_G,X_G'} d_{\bm\theta,X_Y}(X_G,X_G'),    
\end{align}

in which $d_{\bm\theta,X_Y}(X_G,X_G') =\big|P(\widehat{Y}=y'|X_Y,X_G)-P(\widehat{Y}=y'|X_Y,X_G')\big|$ and $\hat{Y}$ is the prediction of the model $\bm\theta$.
The equality holds when $d_{\bm\theta, X_Y}\tteq 0$ always regardless of $X_Y$. 
\end{theorem}
\vspace{-0.25em}
}
The proof of the theorem is in \cref{sec:proof}.
Note that when we take the maximum over $(y, y')$ on both sides of the inequality in \cref{thm1}, the left-hand side of the inequality becomes identical to DEO (\cref{eq:deo}). Therefore, the theorem implies that DEO is upper bounded by the maximum of $d_{\bm\theta,X_Y}(X_G,X_G')$ (in which the maximum is over $X_G$, $X_G'$, $y$, $y'$), which measures the sensitivity of the model with respect to $G$. In other words, the theorem shows that when a counterfactually fair model is sensitive to $X_G$ (\ie, when $\max d_{\bm\theta, X_Y}(X_G, X'_G)$ is large), the model may result in having high DEO in the worst-case. 



\cref{thm1} elucidates why CF-aware methods in \cref{table:method-comparison} often fail to mitigate DEO despite significant improvements in CD. Namely, if the attribute $G$ assumed in \cref{thm1} exists on CelebA and LFW, DEO for the classifiers trained by CF-aware methods can worsen depending on their robustness to $G$. This will be empirically demonstrated using  ``hair length'' as $G$ in \cref{subsec:G_analysis}, together with the results using a controllable synthetic dataset. Furthermore, \cref{thm1} suggests that we can re-establish the relationship between two notions by making counterfactually fair classifiers non-sensitive to $G$. In the following section, we introduce a method to promote a classifier not to depend on $G$ while achieving CF.



\section{Empirical analyses on the effect of $G$ to CF and GF}
\label{sec:G_analysis}

\subsection{\methodnamefull (\oursmethod)}
\label{subsec:ours}
Motivated by \cref{thm1}, we propose a baseline fair-training method to achieve both CF and GF.
Conceptually, if we can reduce the dependency between the latent attribute $G$ described in Theorem \ref{thm1} and the prediction of a CF-aware trained model, we can expect that the model will achieve CF and GF simultaneously.
Therefore, we improve the CF-aware method, CP  \cite{russell2017worlds} (best-performing in \cref{table:method-comparison}), such that the dependency to the attribute $G$ is reduced. We first describe the CP regularization (which is used along with the cross-entropy loss) for given counterfactual samples $\mathcal{D'} = \{x_{i,A\leftarrow a_i'}\}_{i=1}^N$ corresponding to the original training dataset $\mathcal{D}$:
{
\vspace{-0.5em}
\begin{align}
\label{eq:cp}
\mathcal{L}_{\text{CP}}&(\theta,\mathcal{D} \cup\mathcal{D'}) := \frac{1}{N} \sum_{i=1}^N \lVert f(\bm\theta, x_i) - f(\bm\theta, x_{i,A\leftarrow a_i'})\rVert_2^2,
\end{align}
\vspace{-1.25em}
}

in which $f(\bm\theta, x)$ is a representation vector of input $x$ produced by a classifier $\bm\theta$, such as logit or feature vector. Note that the images $x$ and $x_{A\leftarrow a'}$ differ only in their components corresponding to the sensitive attribute $A$ and the attributes caused by the sensitive attribute $A$. Hence, although the CP regularization works well for achieving CF, it does not ensure the model does not rely on the attribute $G$, potentially leading to worse DEO as argued in the previous section.  

Recent studies \cite{jung2021mfd, song2020denoising, zi2021revisiting} have shown that the robustness of a teacher model can be transferred into a student model through knowledge distillation (KD) \cite{hinton2015distilling}. To that end, we first assume a teacher model that is robust to the attribute $G$ is available. Then, our idea is to apply both KD and CP regularization to train our student model, which leads to a simple yet effective approach, dubbed as \methodnamefull (\oursmethod). Specifically, \oursmethod employs averaged representation vectors of original and counterfactual samples extracted by the teacher model $\bm\theta^T$ as target vectors. Then, representation vectors of both samples from the student model $\bm\theta$ are enforced to follow the target vectors. Namely, the distillation term of \oursmethod is defined as follows:
{
\begin{align}
\label{eq:ckd}
 \mathcal{L}&_{\oursmethod}(\bm\theta,\mathcal{D}\cup\mathcal{D}') := \nonumber \frac{1}{2N}\sum_i^N \bigg( \lVert f(\bm\theta,x_i)-f^T_i \rVert_2^2 + \lVert f(\bm\theta, x_{i, A\leftarrow a_i'})-f^T_i \rVert_2^2\bigg), \nonumber\\
&\text{in which}  \ \ f^T_i = \frac{1}{2} \big(f(\bm\theta^T, x_i) + f(\bm\theta^T, x_{i, A\leftarrow a_i'})\big)\ \  \text{is the target vector for the $i$-th pair.}
\end{align}
}
Note that our distillation terms have both effects of KD and CP by promoting both representations of original and counterfactual samples to be aligned with the target vectors $f_i^T$ produced by the teacher model. Therefore, based on \cref{thm1}, we can deduce that the CKD regularization encourages the model to achieve both CF (by the CP effect) and EO-based GF (by the KD effect that distills the robustness of the teacher with respect to the attribute $G$).
In addition, we optionally incorporate CP regularization into our objective to further mitigate CD. The final objective of 
our method (which we again dub as CKD for brevity)
is as follows:
\begin{align}
\label{eq:ckd_full}
\underset{\bm\theta}{\min}\ &\mathcal{L}_{\text{CE}}(\bm\theta,\mathcal{D}) +\mu\mathcal{L}_{\text{\oursmethod}} (\bm\theta, \mathcal{D}\cup\mathcal{D}')+ \lambda\mathcal{L}_{\text{CP}} (\bm\theta, \mathcal{D}\cup \mathcal{D}'),
\end{align}
in which $\mu$ and $\lambda$ are controllable hyperparameters for the CKD and CP regularization, respectively.  

While we assumed above the availability of a teacher model that is robust to the attribute $G$, 
obtaining such a model could be challenging in practice. Empirically, we observe that vanilla-trained models (referred to as ``Scratch'' models) less depend on the attribute $G$ than CP-trained ones (see \cref{fig:cifar-alpha} and \cref{table:CD_G} for more details). We presume that this is because the attributes $A$ and $G$ behave as ``shortcut'' features for classifying the class attribute $Y$, \ie, they are easy-to-learn discriminatory features. As observed by \citet{scimeca2022wcst-ml}, making a model blind to a certain shortcut feature causes it to rely more heavily on the other shortcut features. In our case, CP-trained models are trained to be invariant to the sensitive attribute $A$, resulting in a greater dependence on the attribute $G$ compared to the Scratch models. Thus, unless otherwise specified, we will assume the vanilla-trained model is relatively robust to the attribute $G$ since it would mostly rely on the sensitive attribute $A$, hence, we use it as the teacher model.

\subsection{Impact of robustness to $G$ on CF and GF}
\label{subsec:G_analysis}
We empirically validate our theoretical result and \oursmethod on both a newly introduced synthetic dataset (CIFAR-10B) and a real-world dataset (CelebA) by analyzing CF, GF, and the robustness with respect to the attribute $G$ described in Theorem \ref{thm1}. We thus introduce a new metric for the robustness to the attribute $G$, the rate of flipped predictions (\textbf{RFP}) :

{\small
\vspace{-1.5em}
\begin{align}
\text{RFP}\triangleq\mathbb{E}_{x,x'} \big[P\big(\mathbbm{1} \{\widehat{Y} \neq \widehat{Y}'\}|x,x'\big)\big]. \label{eq:CD-G}
\end{align}
}
in which $x$ is an original image, $x'$ is its corresponding image with the attribute $G$ flipped. $\widehat{Y}$ and $\widehat{Y}'$ refer to the predicted label by the trained model given $x$ and $x'$, respectively. RFP quantifies the amount of flipped predictions when the attribute $G$ is altered. For example, if a model shows the same prediction after changing the attribute  $G$, its RFP becomes 0\%.

\paragraph{CIFAR-10B, a controllable synthetic dataset.} We construct the CIFAR-10B dataset, where we can perfectly control the degree of bias with respect to the attribute $G$ while the target label is biased towards the sensitive attribute $A$. We make binary class labels from the 10 classes of CIFAR-10 (0-4 and 5-9 classes). We set the attributes $A$ and $G$ in \cref{thm1} with the presence of Gaussian and Contrast noise, respectively. We also set a fixed ratio of 0.8 and a controllable ratio $\alpha$, which represent skewnesses among ($Y$, $A$) and ($A$, $G$), respectively; the former ratio is the spurious correlation between $Y$ and $A$, and the latter one is the correlation between $A$ and $G$. We then construct the CIFAR-10B dataset by randomly injecting Gaussian or Contrast noise to each CIFAR-10 image at given ratios, as illustrated in \cref{fig:CIFAR}. Unless otherwise noted, we set $\alpha$ as 0.8.

\begin{figure}[t!]
\centering
\includegraphics[width=0.98\columnwidth]{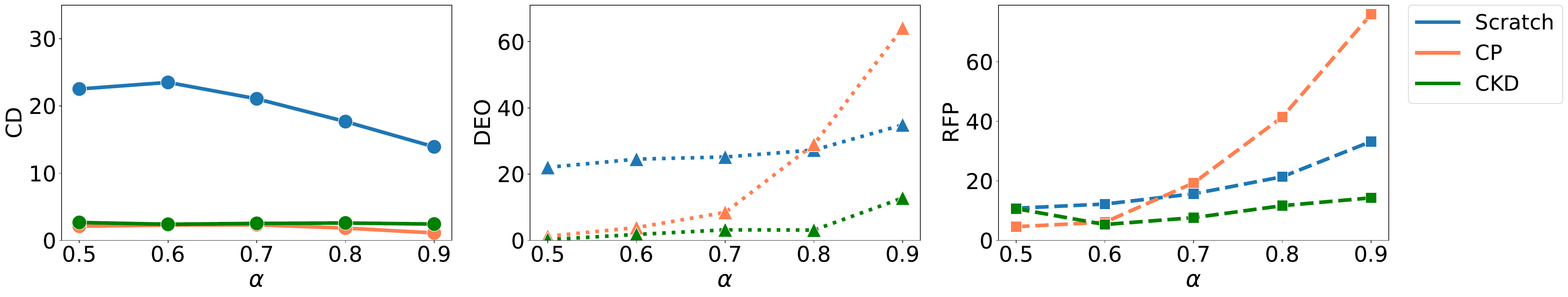}
\caption{\small {\bf Impact of the correlation of $G$ and $A$.} $\alpha$ indicates how $A$ and $G$ are correlated on CIFAR-10B.}
\label{fig:cifar-alpha}
\vspace{-.75em}
\end{figure}

We train models with Scratch, CP, and CKD on CIFAR-10B by adjusting $\alpha$ from 0.5 (\ie, $A$ and $G$ are decorrelated) to 0.9 at intervals of 0.1.
\cref{fig:cifar-alpha} shows CD, DEO, and RFP metric values for each method. The figures indicate that while CP and our CKD consistently achieve CF, CP fails to meet GF as $\alpha$ increases, potentially due to higher RFP. Furthermore, RFP of Scratch is lower than that of CP when $\alpha$ is greater than 0.7. This empirically justifies the use of vanilla-trained models as teacher models robust to $G$. By using these teacher models, CKD significantly improves DEO regardless of the value of $\alpha$ by maintaining the robustness to $G$, \ie, low RFP, supporting the result of \cref{thm1}.

\begin{wraptable}{r}{0.4\textwidth}
\vspace{-1.5em}
\caption{\small {\bf Impact of $G$ on CelebA.} We assume ``hair length'' as $G$ and manipulate the hair length of test images. CD, DEO, and RFP are measured on \oursceleb, CelebA, and hair-edited CelebA, respectively.}
\label{table:CD_G}
\centering
\small
\vspace{.3em}
\begin{tabular}{lccc}
\toprule
                                    & \multicolumn{3}{c}{CelebA}                     \\
Method & CD $\downarrow$ & DEO $\downarrow$ & RFP $\downarrow$  \\
\midrule
\midrule
Scratch& 10.26 & 47.10 & 15.27  \\
\midrule
CP & \textbf{2.53}& 51.01  & 20.37 \\
\oursmethod  & 4.44 & \textbf{13.23} & \textbf{10.85} \\
\bottomrule
\end{tabular}
\vskip -0.2in
\end{wraptable}
\paragraph{Impact of $G$ manipulation on CelebA.} 
We assume ``hair length'' as $G$ for facial image datasets, \eg, CelebA because the hair length $G$ can be highly correlated with, but not caused by, the sex $A$. To compute RFP for the hair length attribute, we manipulate the hair length of CelebA test images using SDEdit \cite{meng2021sdedit}. More details and generated examples can be found in \cref{appendix:CD-G}. Using the hair length-edited images, we report RFP in \cref{table:CD_G}, together with DEO and CD. The results share the same trend as the CIFAR-10B results, \ie, CP shows worse DEO and RFP than Scratch but better CD, whereas CKD shows the best DEO and RFP, despite a slight increase of CD.

\subsection{Impact of the robustness to $G$ of the teacher model on \oursmethod}
\label{subsec:teacher_analysis}
\begin{wraptable}{r}{0.45\textwidth}
\small
\vspace{-1.0em}
\caption{\small {\bf Impact of robustness to $G$ of the teacher model.}
$\bm\theta^T_\text{CP}$, $\bm\theta^T_\text{CKD}$, and $\bm\theta^T_\text{Scratch}$ are CP, CKD, and Scratch teacher model. $\bm\theta^T_\text{De-biased}$ is a Scratch model trained on a perfectly de-biased training dataset ($\alpha=0.5$). RFP$^T$ denotes how a teacher is biased towards $G$. CD, DEO, RFP are metrics for evaluating CF, GF, and bias towards $G$, respectively. Results are measured on CIFAR-10B with $\alpha=0.8$}
\label{table:change-teacher}
\vspace{.3em}
\resizebox{\linewidth}{!}{
\setlength{\tabcolsep}{3pt}
\begin{tabular}{lccccc}
\toprule
Method & RFP$^T$ $\downarrow$ & Acc $\uparrow$  & CD $\downarrow$ & DEO $\downarrow$  & RFP $\downarrow$ \\
\midrule
\midrule
\oursmethod w/ $\bm\theta^T_\text{CP}$ & {41.46} & 76.15 & 3.59 & 12.65  & 18.08 \\ 
\oursmethod w/ $\bm\theta^T_\text{Scratch}$ & {21.38} & 78.49 & {2.85} & {7.30}  & {11.66} \\ 
\oursmethod w/ $\bm\theta^T_\text{CKD}$ & \second{11.66} & 78.39 & \best{2.33} & \second{4.89} & \second{5.30} \\ \midrule
\oursmethod w/ $\bm\theta^T_\text{De-biased}$ & \best{10.81} & 77.17 & \second{2.79}  & \best{4.01} & \best{4.67}  \\
\bottomrule
\end{tabular}
}
\end{wraptable}
Our \oursmethod requires a robust teacher model with respect to the attribute $G$ to distill the robustness to the target model. To analyze the impact of the robustness of the teacher model, we compare various teacher models with different dependencies on the attribute  $G$ using CIFAR-10B.
We consider four teacher models, ordered by robustness to the attribute $G$: CP ($\bm\theta^T_\text{CP}$),  Scratch ($\bm\theta^T_\text{Scratch}$), and CKD model with a Scratch teacher ($\bm\theta^T_\text{CKD}$), and a de-biased model trained on CIFAR-10B balanced for $G$, \ie, $\alpha=0.5$, ($\bm\theta^T_\text{De-biased}$). Using these teacher models, we report DEO, CD, and RFP of CKDs on the CIFAR-10B dataset in \cref{table:change-teacher}. We observe that the degree of robustness to the attribute $G$ of the teacher model (\ie, RFP$^T$) highly correlates to DEO. It is because as the teacher model becomes more robust to $G$, RFP of the target model gets lower, finally leading to a lower DEO while maintaining fair CD. Namely, these results support our theoretical result again.
\section{Full comparisons of fair-training methods on image classification}
\label{sec:exp}
Finally, we evaluate the existing fair-training methods focusing on group fairness (GF) and counterfactual fairness (CF) on CelebA and LFW, together with CIFAR-10B for image classification tasks.
We emphasize that only \oursceleb and \ourslfw have counterfactual images of the real-world images; hence, we measure a CF metric, \ie,  Counterfactual Disparity (CD) (\cref{eq:cd}), using our datasets.
Along with the CF-aware methods, such as CP \cite{russell2017worlds} and \oursmethod, we report the GF-aware methods including SS \cite{idrissi2022simple}, RW \cite{kamiran2012data}, COV \cite{zafar2017fairness}, MFD \cite{jung2021mfd}, and LBC \cite{jiang2020identifying}.
In addition, we report the naive combinations of GF-aware and CF-aware methods, \eg, training GF-aware method with the augmented training dataset with counterfactual images generated by IP2P \cite{brooks2023instructpix2pix} (denoted as ``+aug'') and combinations of the GF-aware methods and the CP regularization (\cref{eq:cp}) (denoted as ``+CP'').
The hyperparameters for all methods besides the GF-aware methods are selected using the same protocol in \cref{sec:pre-exp}, and ones for the GF-aware methods are chosen based on DEO using the same lower bound of the accuracy. Implementation details are provided in \cref{appendix:baselines-details}. 

\begin{table}[t]
\small
\centering
\caption{\small \textbf{Evaluation of GF and CF of fair-training for image classification}. The details are the same as \cref{table:method-comparison}. ``Scratch'' denotes a model trained without considering the notion of fairness through the vanilla cross-entropy loss. ``+aug'' denotes counterfactual (CTF) image augmentation described in \cref{sec:pre-exp}. If a model performs worse than the Scrath model on CD/DEO, we highlight the numbers in \worse{red}. The best performance is highlighted in \colorbox{bestcolor}{\textbf{orange}}, and the second-best performance is highlighted in \colorbox{secondcolor}{\textbf{grey}}.}
\setlength{\tabcolsep}{4.5pt}
\label{table:method-comparison-main}
\begin{tabular}{lccccccccccc}
\toprule
& \multicolumn{3}{c}{CIFAR-10B ($\alpha$=0.8)} && \multicolumn{3}{c}{CelebA (and \oursceleb)} && \multicolumn{3}{c}{LFW (and \ourslfw)} \\
Method         & Acc $\uparrow$  & CD $\downarrow$ & DEO $\downarrow$ && Acc $\uparrow$ & CD $\downarrow$ & DEO $\downarrow$ && Acc $\uparrow$  & CD $\downarrow$ & DEO $\downarrow$\\
\midrule
Scratch        & 78.01 & 17.90 & 27.46 && 95.53   & 10.26  & 47.10    && 90.85  & 18.06 & 7.66   \\
\midrule
SS \cite{idrissi2022simple} & 74.77 & 16.42 & 25.73 && 95.44 & 9.13 & 42.95 && 90.43 & \worse{18.19} & 6.75 \\
RW \cite{kamiran2012data} & 76.53 & 12.15 & 18.94 && 95.16 & 5.50 & 24.21 && 90.87 & \worse{18.68} & 6.92 \\
COV \cite{zafar2017fairness} & 79.03 & 13.90 & 24.05 && 94.42 & 7.72 & 34.04 && 90.85 & 16.43 & 6.99 \\
MFD \cite{jung2021mfd} & 76.84 & 12.24 & 15.39 && 94.37   & 4.61   & 19.00   && 90.47  & 16.07 & 2.15 \\
LBC \cite{jiang2020identifying} & 76.16 & 15.01 & 17.12 && 94.92   & 6.24   & 22.61   && 90.71  & 15.76 & 3.56   \\
\midrule
SS+aug & 73.45 & 9.95 & 15.21 && 95.17 & 5.24 & 40.80 && 89.96 & 15.23 & 6.82 \\
RW+aug & 76.15 & 12.93 & 20.94 && 95.13 & 5.34 & 24.63 && 90.76 & \worse{18.63} & 6.71 \\
COV+aug & 76.52 & 8.17 & 15.04 && 94.08 & 8.11 & 29.03 && 90.47 & 13.65 & 6.78 \\
MFD+aug & 77.10 & 11.16 & 14.79 && 93.78 & 3.87 & 14.36 && 89.90 & \worse{19.36} & 2.47 \\
LBC+aug & 75.82 & 9.01 & 15.29 && 94.39 & 9.32 & 36.08 && 88.66 & 12.41 & 2.79 \\
\midrule
CP \cite{russell2017worlds} & 75.26 & \best{2.05} & \worse{33.23} && 94.10   & \second{2.53} & \worse{51.01}    && 89.77  & 9.20 & \worse{8.74}  \\ \midrule
SS+CP & 76.54 & 3.14 & \second{9.08} && 94.54 & \best{2.40} & 37.97 && 88.7 & \best{6.13} & 4.26 \\
RW+CP & 75.68 & 8.83 & 13.92 && 95.19 & 4.67 & 25.56 && 90.87 & 15.24 & 6.16 \\
COV+CP & 77.74 & 4.30 & 19.42 && 94.29 & 5.36 & \worse{51.63} && 91.23 & 11.91 & 6.52 \\
MFD +CP      & 76.67 & 10.01 & 13.17 && 93.81   & {3.47}   & 23.31   && 89.39   & 15.15 & \second{1.90}  \\
LBC+CP      & 76.88 & 3.02 & 12.45 && 95.12   & 4.72   & 22.78   && 89.92  & 8.33 & 3.02   \\
\midrule
\midrule
\oursmethod($\lambda=0$)  & 76.32 & 8.59 & 11.23 && 94.12 & 4.31 & \second{14.11} && 90.76 & 12.42 & 2.64 \\
\oursmethod    & 78.49 & \second{2.85} & \best{7.30} && 93.08  & 4.44    & \best{13.23}   && 89.26  & \second{7.94}  & \best{1.88}   \\
\bottomrule
\end{tabular}
\end{table}

\cref{table:method-comparison-main} shows the holistic evaluation of CF and GF for all the methods mentioned above on the three image classification tasks. The table shows four important observations.
First, although CP (a CF-aware method) mostly performs the best on CD, it even shows worse DEO than Scratch. We theoretically and empirically discussed the reason in \cref{sec:theory} and \ref{sec:G_analysis}.
Second, the GF-aware methods are effective in improving DEO but have a minimal impact on CD. This suggests that the faithfulness assumption for SCM may not hold, which will be discussed in more details in \cref{appendix:faithful}. Third, the naive combinations of GF-aware and CF-aware methods exhibit much better CD than using the GF-aware methods alone. Additionally, DEOs achieved by the naively combined methods tend to be improved since their training datasets are balanced over the sensitive attributes by incorporating generated samples into the original training datasets.
Lastly, we found that \oursmethod shows the best DEO for every evaluation dataset. It shows that if we can train a CF-aware model by reducing the dependency on $G$, we can achieve both CF and GF even on the image classification task. We additionally conduct an ablation study on \oursmethod by removing CP (\ie, \oursmethod ($\lambda=0$)) from \cref{eq:ckd_full}. We observe that this ablated version achieves suboptimal performances than \oursmethod. This suggests that adding the CP regularization term to the CKD objective function can be helpful to improve both CD and DEO.

\section{Concluding remarks}
This paper offers carefully crafted benchmark datasets for evaluating the counterfactual fairness (CF) of image classification methods. Since obtaining true counterfactual images is impossible in practice, we employ a high-quality image editing technique to generate counterfactual images of the given images. We construct two facial image benchmarks, \oursceleb and \ourslfw, by carefully filtering out and verifying the generated counterfactual images by human annotators. Our datasets relax the constraints of the impossibility of evaluating CF in image classification. Using our datasets, we also provide theoretical and empirical results showing that CF may not imply GF, contradictory to the studies conducted on tabular datasets. We elucidate this phenomenon by the presence of the third-party attribute highly correlated with, but not caused by, the sensitive attribute. From this finding, we propose a simple baseline method, \oursmethod, to achieve CF and GF simultaneously. Our extensive experimental results on both GF and CF metrics show that when reducing the reliance on the attribute (\eg, by using \oursmethod), improving the CF metric leads to a significant improvement in the GF metric.
By providing our benchmarks and various analyses, we believe that our findings bridge CF and GF in image classification, contributing to the development of fair and robust image recognition systems.

\section*{Acknowledgments}
This work was supported in part by the National Research Foundation of Korea (NRF) grant [No.2021R1A2C2007884] and by Institute of Information \& communications Technology Planning \& Evaluation (IITP) grants [RS-2021-II211343, RS-2021-II212068, RS-2022-II220113, RS-2022-II220959] funded by the Korean government (MSIT). It was also supported by AOARD Grant No. FA2386-23-1-4079, SNU-Naver Hyperscale AI Center, and Hyundai Motor Chung Mong-Koo Foundation.

%
{
\small
\bibliography{references}

@String(CVPR= {IEEE Conf. Comput. Vis. Pattern Recog. (CVPR)})

@String(ICCV= {Int. Conf. Comput. Vis. (ICCV)})

@String(ECCV= {Eur. Conf. Comput. Vis. (ECCV)})

@String(NIPS= {Adv. Neural Inform. Process. Syst. (NeurIPS)})

@String(FAccT = {Conf. Fairness, Accountability and Transparency (FAccT)})

@String(TMM  = {IEEE Trans. Multimedia})

@String(ICLR = {Int. Conf. Learn. Represent. (ICLR)})

@String(AAAI = {Proc. of the AAAI Conf. Artificial Intelligence (AAAI)})

@String(ICML = {Int. Conf. Mach. Learn. (ICML)})

@String(AIS = {Int. Conf. Artificial Intelligence and Statistics (AISTATS)})

@String(AIES = {AAAI/ACM Conf. AI, Ethics, and Society (AIES)})

@String(WACV = {IEEE/CVF Winter Conf. App. Comput. Vis. (WACV)})

@String(CLeaR = {Conf. Causal Learning and Reasoning (CLeaR)})

@String(DMLR = {Journal of Data-centric Machine Learning Research (DMLR)})

@String(MLHC = {Machine Learning for Healthcare Conference (MLHC)})

@inproceedings{pfohl2019counterfactual,
  title={Counterfactual reasoning for fair clinical risk prediction},
  author={Pfohl, Stephen R and Duan, Tony and Ding, Daisy Yi and Shah, Nigam H},
  booktitle=MLHC,
  pages={325--358},
  year={2019},
  organization={PMLR}
}

@article{louizos2017causal,
  title={Causal effect inference with deep latent-variable models},
  author={Louizos, Christos and Shalit, Uri and Mooij, Joris M and Sontag, David and Zemel, Richard and Welling, Max},
  journal=NIPS,
  volume={30},
  year={2017}
}

@article{russell2017worlds,
  title={When worlds collide: integrating different counterfactual assumptions in fairness},
  author={Russell, Chris and Kusner, Matt J and Loftus, Joshua and Silva, Ricardo},
  journal=NIPS,
  volume={30},
  year={2017}
}

@inproceedings{rosenblatt2023counterfactual,
  title={Counterfactual fairness is basically demographic parity},
  author={Rosenblatt, Lucas and Witter, R Teal},
  booktitle=AAAI,
  volume={37},
  number={12},
  pages={14461--14469},
  year={2023}
}

@article{pearl2000models,
  title={Models, reasoning and inference},
  author={Pearl, Judea and others},
  journal={Cambridge, UK: CambridgeUniversityPress},
  volume={19},
  number={2},
  pages={3},
  year={2000}
}

@InProceedings{Huang2007a,
  author =    {Gary B. Huang and Vidit Jain and Erik Learned-Miller},
  title =     {Unsupervised Joint Alignment of Complex Images},
  booktitle = ICCV,
  year =      {2007}
}

@inproceedings{peychev2022latent,
  title={Latent space smoothing for individually fair representations},
  author={Peychev, Momchil and Ruoss, Anian and Balunovi{\'c}, Mislav and Baader, Maximilian and Vechev, Martin},
  booktitle=ECCV,
  pages={535--554},
  year={2022},
  organization={Springer}
}

@article{yurochkin2020sensei,
  title={Sensei: Sensitive set invariance for enforcing individual fairness},
  author={Yurochkin, Mikhail and Sun, Yuekai},
  journal={arXiv preprint arXiv:2006.14168},
  year={2020}
}

@inproceedings{brooks2023instructpix2pix,
  title={Instructpix2pix: Learning to follow image editing instructions},
  author={Brooks, Tim and Holynski, Aleksander and Efros, Alexei A},
  booktitle=CVPR,
  pages={18392--18402},
  year={2023}
}

@article{luccioni2023stable,
  title={Stable bias: Analyzing societal representations in diffusion models},
  author={Luccioni, Alexandra Sasha and Akiki, Christopher and Mitchell, Margaret and Jernite, Yacine},
  journal={arXiv preprint arXiv:2303.11408},
  year={2023}
}

@article{meng2021sdedit,
  title={Sdedit: Guided image synthesis and editing with stochastic differential equations},
  author={Meng, Chenlin and He, Yutong and Song, Yang and Song, Jiaming and Wu, Jiajun and Zhu, Jun-Yan and Ermon, Stefano},
  journal={arXiv preprint arXiv:2108.01073},
  year={2021}
}

@article{kingma2013auto,
  title={Auto-encoding variational bayes},
  author={Kingma, Diederik P and Welling, Max},
  journal={arXiv preprint arXiv:1312.6114},
  year={2013}
}

@article{goodfellow2014generative,
  title={Generative adversarial nets},
  author={Goodfellow, Ian and Pouget-Abadie, Jean and Mirza, Mehdi and Xu, Bing and Warde-Farley, David and Ozair, Sherjil and Courville, Aaron and Bengio, Yoshua},
  journal=NIPS,
  volume={27},
  year={2014}
}

@article{song2020denoising,
  title={Denoising diffusion implicit models},
  author={Song, Jiaming and Meng, Chenlin and Ermon, Stefano},
  journal={arXiv preprint arXiv:2010.02502},
  year={2020}
}

@inproceedings{wu2023consistency,
  title={Consistency and accuracy of Celeba attribute values},
  author={Wu, Haiyu and Bezold, Grace and G{\"u}nther, Manuel and Boult, Terrance and King, Michael C and Bowyer, Kevin W},
  booktitle=CVPR,
  pages={3257--3265},
  year={2023}
}

@article{kingma2018glow,
  title={Glow: Generative flow with invertible 1x1 convolutions},
  author={Kingma, Durk P and Dhariwal, Prafulla},
  journal=NIPS,
  volume={31},
  year={2018}
}

@inproceedings{liang2023benchmarking,
  title={Benchmarking Algorithmic Bias in Face Recognition: An Experimental Approach Using Synthetic Faces and Human Evaluation},
  author={Liang, Hao and Perona, Pietro and Balakrishnan, Guha},
  booktitle=ICCV,
  pages={4977--4987},
  year={2023}
}

@inproceedings{goldblum2020adversarially,
  title={Adversarially robust distillation},
  author={Goldblum, Micah and Fowl, Liam and Feizi, Soheil and Goldstein, Tom},
  booktitle=AAAI,
  volume={34},
  number={04},
  pages={3996--4003},
  year={2020}
}

@inproceedings{dash2022evaluating,
  title={Evaluating and mitigating bias in image classifiers: A causal perspective using counterfactuals},
  author={Dash, Saloni and Balasubramanian, Vineeth N and Sharma, Amit},
  booktitle=WACV,
  pages={915--924},
  year={2022}
}

@article{kocaoglu2017causalgan,
  title={Causalgan: Learning causal implicit generative models with adversarial training},
  author={Kocaoglu, Murat and Snyder, Christopher and Dimakis, Alexandros G and Vishwanath, Sriram},
  journal={arXiv preprint arXiv:1709.02023},
  year={2017}
}

@inproceedings{anthis2023causal,
  title={Causal Context Connects Counterfactual Fairness to Robust Prediction and Group Fairness},
  author={Anthis, Jacy Reese and Veitch, Victor},
  booktitle=NIPS,
  year={2023}
}

@inproceedings{zi2021revisiting,
  title={Revisiting adversarial robustness distillation: Robust soft labels make student better},
  author={Zi, Bojia and Zhao, Shihao and Ma, Xingjun and Jiang, Yu-Gang},
  booktitle=ICCV,
  pages={16443--16452},
  year={2021}
}

@inproceedings{idrissi2022simple,
  title={Simple data balancing achieves competitive worst-group-accuracy},
  author={Idrissi, Badr Youbi and Arjovsky, Martin and Pezeshki, Mohammad and Lopez-Paz, David},
  booktitle=CLeaR,
  pages={336--351},
  year={2022},
  organization={PMLR}
}

@inproceedings{kim2021counterfactual,
  title={Counterfactual fairness with disentangled causal effect variational autoencoder},
  author={Kim, Hyemi and Shin, Seungjae and Jang, JoonHo and Song, Kyungwoo and Joo, Weonyoung and Kang, Wanmo and Moon, Il-Chul},
  booktitle=AAAI,
  volume={35},
  number={9},
  pages={8128--8136},
  year={2021}
}

@inproceedings{ramaswamy2021fair,
  title={Fair attribute classification through latent space de-biasing},
  author={Ramaswamy, Vikram V and Kim, Sunnie SY and Russakovsky, Olga},
  booktitle=CVPR,
  pages={9301--9310},
  year={2021}
}

@inproceedings{xu2018fairgan,
  title={Fairgan: Fairness-aware generative adversarial networks},
  author={Xu, Depeng and Yuan, Shuhan and Zhang, Lu and Wu, Xintao},
  booktitle={2018 IEEE International Conference on Big Data (Big Data)},
  pages={570--575},
  year={2018},
  organization={IEEE}
}

@inproceedings{zhang2022fairness,
  title={Fairness-aware contrastive learning with partially annotated sensitive attributes},
  author={Zhang, Fengda and Kuang, Kun and Chen, Long and Liu, Yuxuan and Wu, Chao and Xiao, Jun},
  booktitle=ICLR,
  year={2022}
}

@inproceedings{d2024improving,
  title={Improving Fairness using Vision-Language Driven Image Augmentation},
  author={D'Inc{\`a}, Moreno and Tzelepis, Christos and Patras, Ioannis and Sebe, Nicu},
  booktitle=WACV,
  pages={4695--4704},
  year={2024}
}

@article{nguyen2016hirability,
  title={Hirability in the wild: Analysis of online conversational video resumes},
  author={Nguyen, Laurent Son and Gatica-Perez, Daniel},
  journal=TMM,
  volume={18},
  number={7},
  pages={1422--1437},
  year={2016},
  publisher={IEEE}
}

@inproceedings{buolamwini2018gender,
  title={Gender shades: Intersectional accuracy disparities in commercial gender classification},
  author={Buolamwini, Joy and Gebru, Timnit},
  booktitle=FAccT,
  pages={77--91},
  year={2018},
  organization={PMLR}
}

@inproceedings{wang2019racial,
  title={Racial faces in the wild: Reducing racial bias by information maximization adaptation network},
  author={Wang, Mei and Deng, Weihong and Hu, Jiani and Tao, Xunqiang and Huang, Yaohai},
  booktitle=ICCV,
  pages={692--702},
  year={2019}
}

@article{kamiran2012data,
  title={Data preprocessing techniques for classification without discrimination},
  author={Kamiran, Faisal and Calders, Toon},
  journal={Knowledge and Information Systems (KAIS)},
  volume={33},
  number={1},
  pages={1--33},
  year={2012},
  publisher={Springer}
}

@inproceedings{zafar2017fairness,
  title={Fairness constraints: Mechanisms for fair classification},
  author={Zafar, Muhammad Bilal and Valera, Isabel and Rogriguez, Manuel Gomez and Gummadi, Krishna P},
  booktitle=AIS,
  pages={962--970},
  year={2017},
  organization={PMLR}
}

@inproceedings{jung2021mfd,
  title={Fair Feature Distillation for Visual Recognition},
  author={Jung, Sangwon and Lee, Donggyu and Park, Taeeon and Moon, Taesup},
  booktitle=CVPR,
  pages={12115--12124},
  year={2021}
}

@inproceedings{jiang2020identifying,
  title={Identifying and correcting label bias in machine learning},
  author={Jiang, Heinrich and Nachum, Ofir},
  booktitle=AIS,
  pages={702--712},
  year={2020},
  organization={PMLR}
}

@inproceedings{celeba,
  title={Deep learning face attributes in the wild},
  author={Liu, Ziwei and Luo, Ping and Wang, Xiaogang and Tang, Xiaoou},
  booktitle=ICCV,
  pages={3730--3738},
  year={2015}
}

@inproceedings{dutta2020there,
  title={Is there a trade-off between fairness and accuracy? {A} perspective using mismatched hypothesis testing},
  author={Dutta, Sanghamitra and Wei, Dennis and Yueksel, Hazar and Chen, Pin-Yu and Liu, Sijia and Varshney, Kush},
  booktitle=ICML,
  pages={2803--2813},
  year={2020},
  organization={PMLR}
}

@inproceedings{hardt2016equality,
  title={Equality of opportunity in supervised learning},
  author={Hardt, Moritz and Price, Eric and Srebro, Nati},
  booktitle=NIPS,
  volume = {29},
  year={2016}
}

@incollection{kennaway2020causation,
  title={When causation does not imply correlation: Robust violations of the faithfulness axiom},
  author={Kennaway, Richard},
  booktitle={The Interdisciplinary Handbook of Perceptual Control Theory},
  pages={49--72},
  year={2020},
  publisher={Elsevier}
}

@inproceedings{kusner2017counterfactual,
  title={Counterfactual fairness},
  author={Kusner, Matt J and Loftus, Joshua R and Russell, Chris and Silva, Ricardo},
  booktitle=NIPS,
  year={2017}
}

@article{hinton2015distilling,
  title={Distilling the knowledge in a neural network},
  author={Hinton, Geoffrey and Vinyals, Oriol and Dean, Jeff},
  journal={arXiv preprint arXiv:1503.02531},
  year={2015}
}

@article{objective_biased,
 author  = {Harlan, Elisa and Schnuck, Oliver},
 date    = {2021-02-16},
 title   = {Objective or biased – The questionable use of Artificial Intelligence in job applications},
 journal = {bayerischer rundfunk},
 url     = {https://interaktiv.br.de/ki-bewerbung/en/},
 date    = {2021-02-16},
year={2021}
}

@inproceedings{garg2019counterfactual,
  title={Counterfactual fairness in text classification through robustness},
  author={Garg, Sahaj and Perot, Vincent and Limtiaco, Nicole and Taly, Ankur and Chi, Ed H and Beutel, Alex},
  booktitle=AIES,
  pages={219--226},
  year={2019}
}

@inproceedings{scimeca2022wcst-ml,
    title={Which Shortcut Cues Will DNNs Choose? A Study from the Parameter-Space Perspective}, 
    author={Luca Scimeca and Seong Joon Oh and Sanghyuk Chun and Michael Poli and Sangdoo Yun},
    year={2022},
    booktitle=ICLR,
}

@article{pinto2024matrix,
  title={The Matrix Reloaded: Towards Counterfactual Group Fairness in Machine Learning},
  author={Pinto, Mariana and Carreiro, Andre V and Madeira, Pedro and Lopez, Alberto and Gamboa, Hugo},
  journal=DMLR,
  year={2024}
}
\bibliographystyle{abbrvnat}
}
\clearpage
\appendix
\section*{Checklist}

The checklist follows the references.  Please
read the checklist guidelines carefully for information on how to answer these
questions.  For each question, change the default \answerTODO{} to \answerYes{},
\answerNo{}, or \answerNA{}.  You are strongly encouraged to include a {\bf
justification to your answer}, either by referencing the appropriate section of
your paper or providing a brief inline description.  For example:
\begin{itemize}
  \item Did you include the license to the code and datasets? \answerYes{See Section~\ref{appendix:datasheet}.}
  \item Did you include the license to the code and datasets? \answerNo{The code and the data are proprietary.}
  \item Did you include the license to the code and datasets? \answerNA{}
\end{itemize}
Please do not modify the questions and only use the provided macros for your
answers.  Note that the Checklist section does not count towards the page
limit.  In your paper, please delete this instructions block and only keep the
Checklist section heading above along with the questions/answers below.

\begin{enumerate}

\item For all authors...
\begin{enumerate}
  \item Do the main claims made in the abstract and introduction accurately reflect the paper's contributions and scope?
    \answerYes{Our claims are reflected accurately.}
  \item Did you describe the limitations of your work?
    \answerYes{See \cref{appendix:limitation}}
  \item Did you discuss any potential negative societal impacts of your work?
    \answerYes{See \cref{appendix:limitation}}
  \item Have you read the ethics review guidelines and ensured that your paper conforms to them?
    \answerYes{}
\end{enumerate}

\item If you are including theoretical results...
\begin{enumerate}
  \item Did you state the full set of assumptions of all theoretical results?
    \answerYes{See \cref{sec:theory}}
	\item Did you include complete proofs of all theoretical results?
    \answerYes{See \cref{sec:proof}}
\end{enumerate}

\item If you ran experiments (e.g. for benchmarks)...
\begin{enumerate}
  \item Did you include the code, data, and instructions needed to reproduce the main experimental results (either in the supplemental material or as a URL)?
    \answerYes{See \cref{appendix:implementation}}
  \item Did you specify all the training details (e.g., data splits, hyperparameters, how they were chosen)?
    \answerYes{See \cref{appendix:implementation}}
	\item Did you report error bars (e.g., with respect to the random seed after running experiments multiple times)?
    \answerYes{See \cref{appendix:subsec_std}}
	\item Did you include the total amount of compute and the type of resources used (e.g., type of GPUs, internal cluster, or cloud provider)?
    \answerYes{See \cref{appendix:subsec_opt}}
\end{enumerate}

\item If you are using existing assets (e.g., code, data, models) or curating/releasing new assets...
\begin{enumerate}
  \item If your work uses existing assets, did you cite the creators?
    \answerYes{We cited \citet{celeba}, \citet{Huang2007a} and \citet{brooks2023instructpix2pix}.}
  \item Did you mention the license of the assets?
    \answerYes{See \cref{appendix:license}}
  \item Did you include any new assets either in the supplemental material or as a URL?
    \answerYes{See \cref{appendix:datasheet}}
  \item Did you discuss whether and how consent was obtained from people whose data you're using/curating?
    \answerNA{We did not create new data but edited existing image benchmarks, so this issue does not apply.}
  \item Did you discuss whether the data you are using/curating contains personally identifiable information or offensive content?
    \answerNA{We did not create new data but edited existing image benchmarks, so this issue does not apply.}
\end{enumerate}

\item If you used crowdsourcing or conducted research with human subjects...
\begin{enumerate}
  \item Did you include the full text of instructions given to participants and screenshots, if applicable?
    \answerYes{See \cref{fig:image-filtering-ui}, \ref{fig:image-filtering-guideline}, and \ref{fig:reliability-checking}.}
  \item Did you describe any potential participant risks, with links to Institutional Review Board (IRB) approvals, if applicable?
    \answerNA{}
  \item Did you include the estimated hourly wage paid to participants and the total amount spent on participant compensation?
    \answerYes{See \cref{appendix:human_evaluation}}
\end{enumerate}

\end{enumerate}
\newpage
\numberwithin{figure}{section}
\numberwithin{table}{section}
\numberwithin{equation}{section}
\section{Proof of \cref{thm1}}\label{sec:proof}
\label{sec:proof}
We start from LHS in equation \ref{eq:theorem}:
\begin{align}
\big|P(\widehat{Y}=y'|A=0,&Y=y)-P(\widehat{Y}=y'|A=1,Y=y)\big| \nonumber \\
=\bigg|\sum_{X_A,X_Y,X_G}P(\widehat{Y}=&y'|X_A,X_Y,X_G,A=0,Y=y)P(X_A,X_Y,X_G|A=0,Y=y) \nonumber \\
&-P(\widehat{Y}=y'|X_A,X_Y,X_G,A=1,Y=y)P(X_A,X_Y,X_G|A=1,Y=y)\bigg| 
\end{align}
\begin{align}
=\bigg|\sum_{X_A,X_Y,X_G}P(\widehat{Y}=y'|X_Y,X_G)P(X_A,&X_Y,X_G|A=0,Y=y)\nonumber\\
&-P(\widehat{Y}=y'|X_Y,X_G)P(X_A,X_Y,X_G|A=1,Y=y)\bigg| \\
=\bigg|\sum_{X_Y,X_G}P(\widehat{Y}=y'|X_Y,X_G)\big(P(X_Y|X_{G}&,A=0,Y=y)P(X_G|A=0,Y=y) \nonumber \\
&-P(X_Y|X_G,A=1,Y=y)P(X_G|A=1,Y=y)\big)\bigg| 
\end{align}
\begin{align}
=\bigg|\sum_{X_Y,X_G}P(\widehat{Y}=y'|X_Y,X_G)\bigg(&P(X_Y|Y=y)P(X_G|A=0,Y=y)-P(X_Y|Y=y)P(X_G|A=1,Y=y)\bigg)\bigg| \\
=\bigg|\sum_{X_Y}P(X_Y|Y=y)\Big(\sum_{X_G}P(\widehat{Y}&=y'|X_Y,X_G)P(X_G|A=0,Y=y)\\
&-\sum_{X_G'}P(\widehat{Y}=y'|X_Y,X_G')P(X_G'|A=1,Y=y)\Big)\bigg|. 
\end{align}
Note the first and third equalities are driven by Bayes' theorem, the second one is from the independence between $\widehat{Y}$ and $X_G,A$ conditioned on $X_Y, X_G$ based on the Markov properties of SCM, and the fourth one is due to the independence between $X_Y$ and $X_G, A$ conditioned on $Y$. We denote a coupling between the two distributions $P(X_G|A=0,Y)$ and $P(X_G'|A=1,Y)$ as $\Pi(X_G,X_G')$, then we have:
\begin{align}
  &   \big|P(\widehat{Y}=y'|A=0,Y=y)-P(\widehat{Y}=y'|A=1,Y=y)\big| \nonumber \\
    & =\bigg|\sum_{X_Y}P(X_Y|Y=y)\Big(\sum_{X_G,X_G'}\Pi(X_G,X_G') \big( P(\widehat{Y}=y'|X_Y,X_G)-P(\widehat{Y}=y'|X_Y,X_G')\big)\Big)\bigg|. \\
& \leq\sum_{X_Y}P(X_Y|Y=y)\bigg(\sum_{X_G,X_G'}\Pi(X_G,X_G') \Big|P(\widehat{Y}=y'|X_Y,X_G)-P(\widehat{Y}=y'|X_Y,X_G')\Big|\bigg) \label{eq:jensen}\\
& =\sum_{X_Y}P(X_Y|Y=y)\sum_{X_G,X_G'}\Pi(X_G,X_G') d_{\bm\theta,X_Y}(X_G,X_G') 
\end{align}
where the sample distance is denoted as $d_{\bm\theta,X_Y}(X_G,X_G') =\big|P(\widehat{Y}=y'|X_Y,X_G)-P(\widehat{Y}=y'|X_Y,X_G')\big|$. The inequality in Equation \ref{eq:jensen} is driven by Jensen's inequality.
\begin{figure}[t!]
\begin{center}
\centerline{\includegraphics[width=0.95\columnwidth]{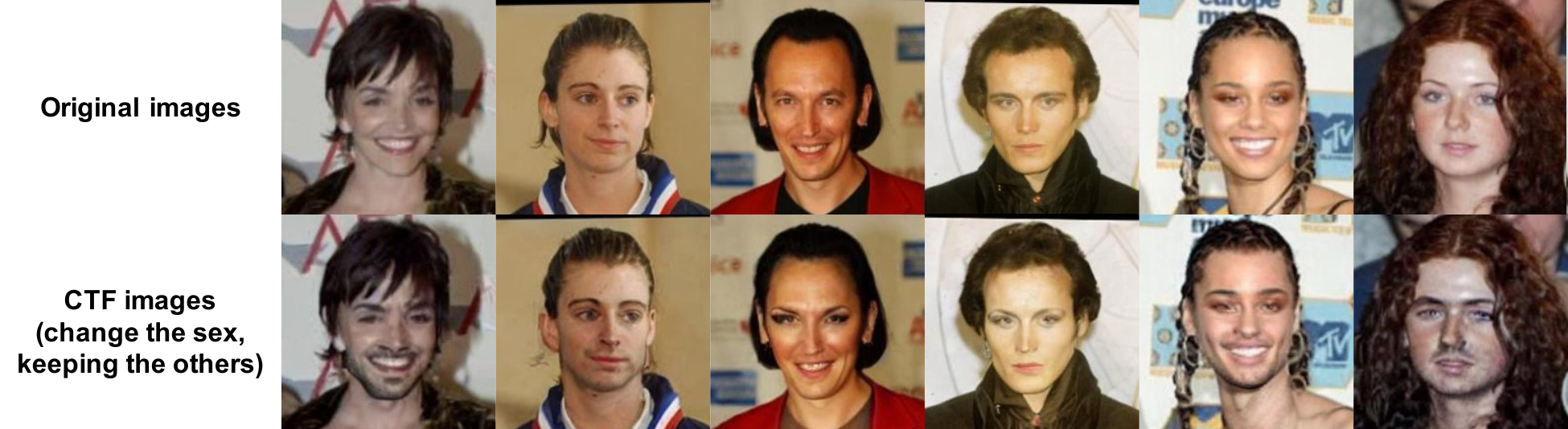}}
\caption{\small {\bf \ourslfw examples}. The counterfactual (CTF) images regarding the ``sex attribute'' are shown. The top row shows the original image, while the bottom row displays the CTF image generated by IP2P.}
\label{fig:ctf_expample_lfw}
\end{center}
\end{figure}

\begin{figure}[t!]
\centering
\includegraphics[width=0.8\columnwidth]{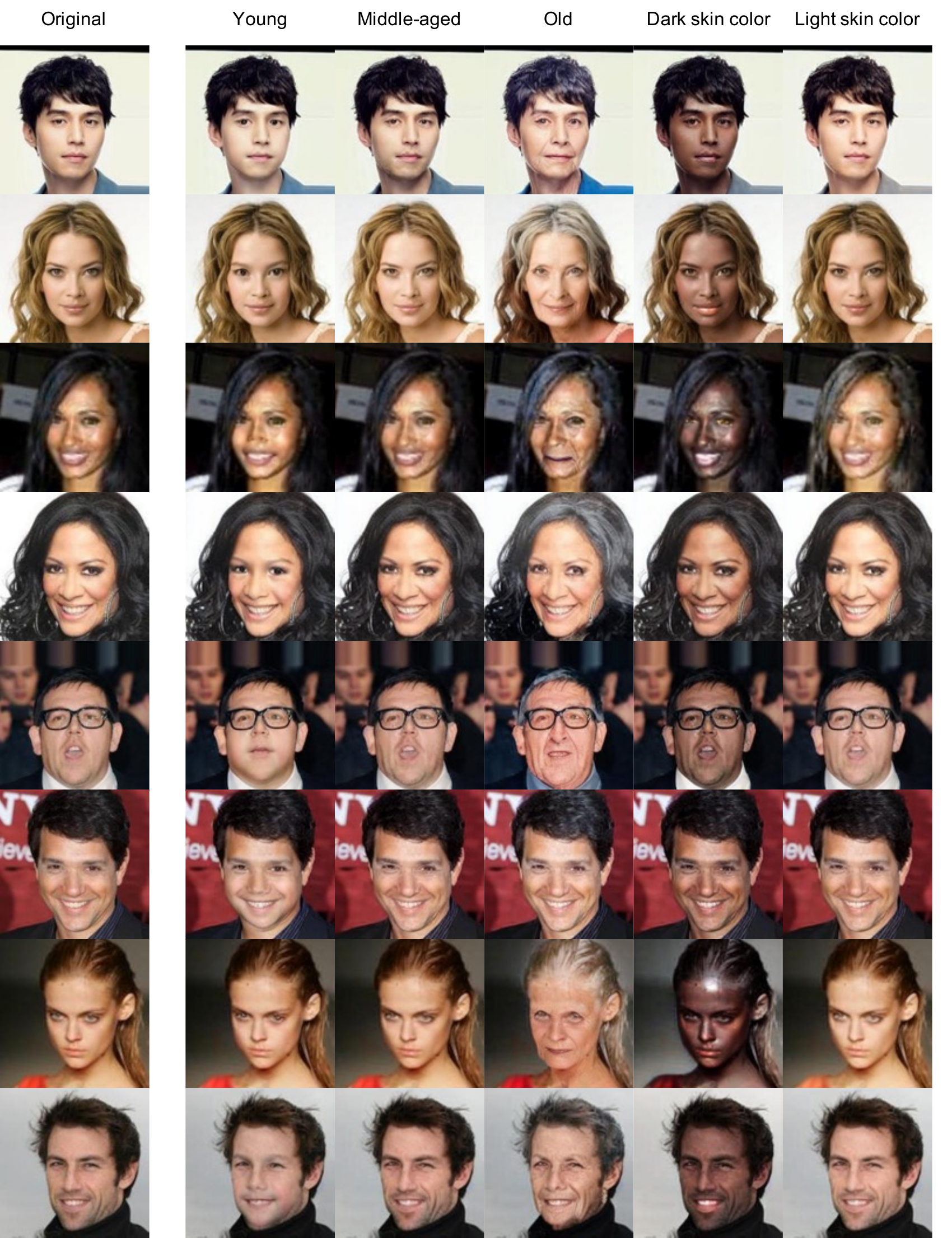}
\caption{\small \textbf{CF examples with other sensitive attributes}. Original and CTF samples are shown when age or skin color is considered as the sensitive attribute.}
\label{fig:ctf_example_multi_celeba}
\end{figure}

\section{Limitations and societal impacts}
\label{appendix:limitation}
While our datasets and analyses reveal the relationship between CF and GF in image classification, we clarify our study's limitations.
First of all, our study uses sex as a sensitive attribute based on visually perceived biological traits.
However, as mentioned in \cref{sec:ctf_contruction}, this simplification does not capture the full spectrum of sexual traits, which is more complex and nuanced.
Therefore, we emphasize again that practitioners should use our data with these considerations in mind; they should not utilize our datasets for gender categorization but rather for investigating the unfairness in terms of CF and GF and enhancing fairness in AI systems. 
Second, our data generation process relies on IP2P to create CTF samples. 
We tried to mitigate the potential bias problem during the data generation process through the sophisticated human filtering process, but our data samples could be affected by the unintended bias of IP2P.
Third, while we assume the structural causal model (SCM) for images as \cref{fig:scm}, specifying an SCM in the real world is often infeasible. This difficulty also makes it challenging to apply some of our experiments, such as analyzing the attribute $G$ or using a robust teacher model to $G$. However, we addressed these challenges to some extent by proposing a systematic method to investigate $G$ (\cref{appendix:subsec_selection_g}) and analyzing the robustness of vanilla teacher models (\cref{subsec:teacher_analysis}). 

Despite the limitations, we believe that our work makes significant contributions through our curated image datasets and extensive analyses to the evolving field that addresses relationships among different fairness notions. 

\section{Datasets: CelebA-CF and LFW-CF}
\label{appendix:our-dataset}
We provide access to our newly created dataset, \ie, CelebA-CF and LFW-CF, through the following link: 
CelebA-CF (\url{https://figshare.com/s/62b6f7f69d0eab9c3c71})
, LFW-CF (\url{https://figshare.com/s/39f2daac58148e10e5fe})

\subsection{Hyperparameters for IP2P image editing}
\label{appendix:subsec_hyp_ip2p} 
We set the resolution of generated images to 256$\times$256 and the denoising step to 50. Furthermore, we set the Image-CFG weight to 1.8 and the Text-CFG weight to 7.5. These two hyperparameters are the guidance scales to control how the generated images closely resemble the input image or are intensely edited. To alter the sensitive attribute of facial images, we use the prompts of ``turn the woman into a man'' for female images and ``turn the man into a woman'' for male images. 


\subsection{Human annotations}
\label{appendix:human_evaluation}
\begin{figure}[t!]
    \centering
    \includegraphics[width=0.7\textwidth]{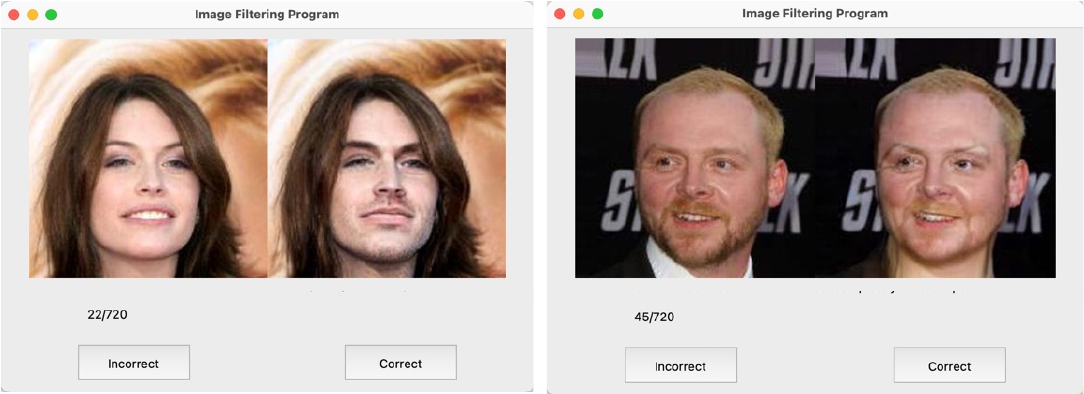}
    \caption{\small User Interface shown to five annotators for image filtering.}
    \label{fig:image-filtering-ui}
\end{figure}

\begin{figure}[t!]
    \centering
    \includegraphics[width=0.9\textwidth]{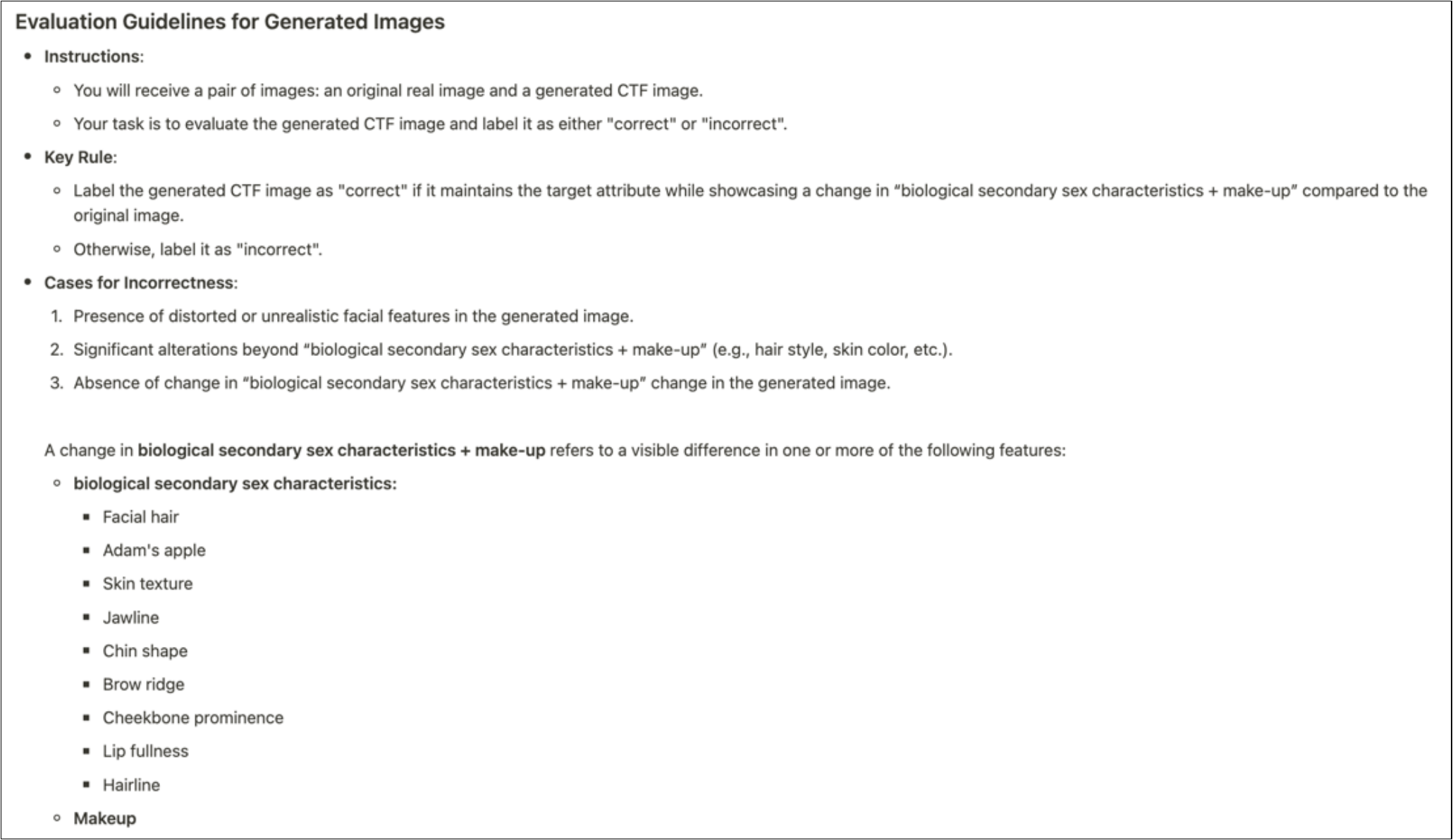}
    \caption{\small The guideline instructions that were given to the five annotators for the image filtering.}
    \label{fig:image-filtering-guideline}
\end{figure}

After generating CTF images using IP2P, we filtered them through five annotators to construct high-quality CTF samples, namely \oursceleb and \ourslfw. Before evaluating the created CTF samples, the guidelines are given to the annotators, as presented in \cref{fig:image-filtering-guideline}. To establish the guidelines, we extracted 20 facial attributes of secondary sex characteristics using Chat-GPT and then, with guidance from experts specialized in fairness, selected 9 key facial attributes (facial hair, Adam's apple, skin texture, jawline, chin shape, brow ridge, cheekbone prominence, lip fullness, and hairline). The guidelines instruct human annotators to filter out counterfactual samples based on these attributes (including considerations for the presence of makeup). Subsequently, given the original image and generated CTF image pair, annotators assess whether the generated image is correctly created based on the instructions. \cref{fig:image-filtering-ui} shows the interfaces of the annotation task for image filtering.

\begin{figure}[t!]
    \centering
    \includegraphics[width=0.7\textwidth]{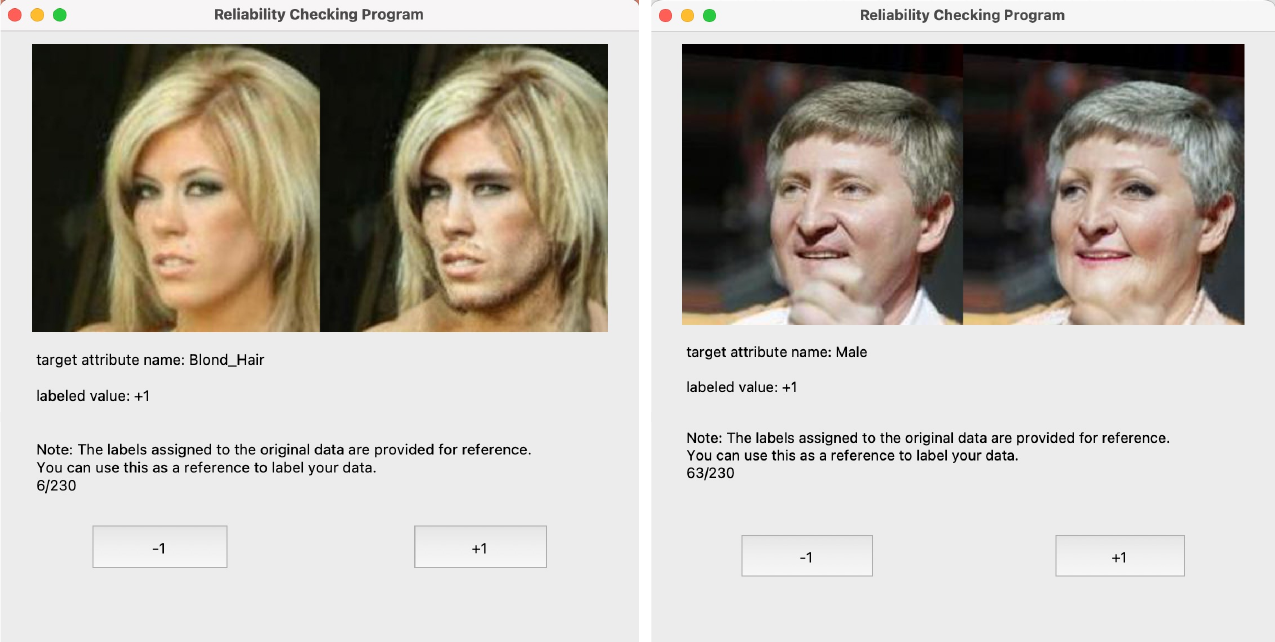}
    \caption{\small User Interface shown to five annotators to evaluate the reliability of our created datasets.}
    \label{fig:reliability-checking}
\end{figure}

We further verify the reliability of our datasets with another five human annotators, different from those who participated in the previous filtering process, and report the result in \cref{table:double-check}. For more objective annotation, we show 8 example images to the annotators before the labeling task, which are randomly sampled the same number of times for each attribute value from test image datasets. Then, the annotators label \oursceleb and \ourslfw for 4 attributes, \ie, ``sex'', ``blond hair'', ``gray hair'', and ``smiling'' in order. Specifically, the annotators evaluate whether the sensitive attribute was correctly altered and the non-sensitive attributes were maintained for a generated image. Similar to the image filtering task, we provide the annotators with the set of 10 sex-related facial attributes for objective and accurate labeling.  \cref{fig:reliability-checking} shows the interfaces of the annotation task for this reliability check. We provide the attribute values originally annotated for the original image datasets on the screen together for annotators to refer to as a guide for their annotating tasks. Although we focus on visually perceived sex traits, we use the terms Male and Female for convenience in the annotation interface.

We note that the wage paid to each participant is 18 USD per hour, resulting in a total expenditure of 360 USD on participant compensation.

\subsection{License information of assets employed in this study}
\label{appendix:license}
\begin{itemize}
    \item CelebA \cite{celeba} was made available for academic research purposes without a formal license. The dataset can be downloaded at \url{https://mmlab.ie.cuhk.edu.hk/projects/CelebA.html}.
    \item LFW \cite{Huang2007a} is publicly available for research purposes. There is also no formal license and further information is reported at \url{https://vis-www.cs.umass.edu/lfw/}.
    \item InstructPix2Pix (IP2P) \cite{brooks2023instructpix2pix} is licensed under the \href{https://mit-license.org}{MIT license} and is available at \url{https://github.com/timothybrooks/instruct-pix2pix}.
    \item IP2P further builds upon stable-diffusion-v1-5 that is released under \href{https://dezgo.com/license}{CreativeML- Open-RAIL-M License}.
\end{itemize}

\subsection{Further information of the new dataset.}
\label{appendix:analysis-dataset}
CelebA-CF and LFW-CF are based on real facial image datasets, such as CelebA and LFW, which include 40 attribute annotations. We note that because the original datasets have an imbalance between some pairs of two attributes, our datasets also possess a different skewness between the sensitive attribute and other attributes. For example, in CelebA and CelebA-CF, most images with the blond hair attribute are female, and most males do not have blond hair. We present the skewness values between the sensitive attribute and other non-sensitive attributes in CelebA-CF in \cref{table:attriabute-imbalance}. Additionally, we displayed the group-specific failure rate identified through the filtering process in \cref{table:failure-rate}.

\begin{table}[t!]
\centering
\small
\caption{\small \textbf{The skewness between the sensitive attribute and other non-sensitive attributes in CelebA-CF.}}
\label{table:attriabute-imbalance}
\setlength{\tabcolsep}{2.0pt}
\vskip 0.15in
\begin{tabular}{cccccccccc}
\toprule
 Attribute & Hair Length & Bangs & Wearing Hat & Brown Hair & Pale Skin & Big Lips & Mouth Slightly Open & Smiling & Wavy Hair \\
\midrule
\midrule
Skewness & 0.798 & 0.907 & 0.974 & 0.820 & 0.980 & 0.815 & 0.265 & 0.335 & 0.833 \\
\bottomrule
\end{tabular}
\vskip -0.1in
\end{table}

\begin{table}[t!]
\centering
\small
\caption{\small \textbf{The failure rate identified through the filtering process.} This table shows the proportion of images filtered out, calculated separately for each sensitive attribute (\eg, female, male), in constructing the CelebA-CF and LFW-CF datasets. The sensitive attributes in the table represent the labels of the original images.}
\label{table:failure-rate}
\vskip 0.15in
\begin{tabular}{ccc}
\toprule
& CelebA-CF & LFW-CF \\
\midrule
\midrule
Female & 0.57 & 0.70 \\
Male & 0.80 & 0.69 \\
\bottomrule
\end{tabular}
\vskip -0.1in
\end{table}

\section{Implementation details}
\label{appendix:implementation}

\subsection{Details on training datasets}
For CelebA, we utilize the official train-validation-test split \cite{celeba}. For LFW, we also use the official train-test split \cite{Huang2007a} and then divide the training data into a training and a validation set, with a ratio of 80:20. 

CIFAR-10B is a modified dataset from CIFAR-10, as described in \cref{subsec:G_analysis}. We modify CIFAR-10 into a binary classification task by dividing the original 10 classes into two classes (classes 0-4 and 5-9). To introduce a fairness issue, we set the sensitive attribute $A$ as the presence of Gaussian noise and skew the dataset by randomly injecting the noise into 20\% and 80\% of the data in class 0 and class 1, respectively. Additionally, we introduce Contrast noise for the attribute $G$. Using the skew-ratio $\alpha$, we create a statistical correlation between $A$ and $G$ by adding the noise into $100\times\alpha$\% of the data samples with $A=1$ and $100\times(1-\alpha)$\% of the data samples with $A=0$. Unless otherwise noted, we set $\alpha$ to 0.8. We partition the dataset into train-validation-test sets with a ratio of 64:16:20, respectively, maintaining consistent values for the two skewness ratios (\ie, skewness between $A$ and $Y$, $G$ and $A$) across all sets during our experiments.

\subsection{Compute Infrastructure and optimization}
\label{appendix:subsec_opt}
Our all experiments including the dataset construction and performance comparison of existing methods and \oursmethod were conducted using AMD Ryzen Threadripper PRO 3975WX CPUs and NVIDIA RTX A5000 GPUs. Our dataset generation was parallelized using 8 GPUs and took 2 days to complete. Training time for models used in the experiments for performance comparison ranges between 12 to 24 hours depending on the dataset and method used.

For CIFAR-10B, we use ResNet56 models with the Adam optimizer for 50 epochs. We set the mini-batch size and learning rate as 128 and 0.001, respectively.
Because the skewness between $G$ and $A$ in CIFAR-10B test datasets varies, we compute the balanced CD over both the target class and the sensitive attribute for the consistent metric.
For CelebA and LFW, we train ResNet18 models with the AdamW optimizer. We use the epoch size of 70 and 50 for each dataset, and set the mini-batch size, learning rate, and weight decay as 128, 0.001, and 1e-4, respectively. We use identical hyperparameters regarding the optimization for all methods. All results are averaged over results from four different random seeds.



\subsection{Implementation details of baselines and CKD}
\label{appendix:baselines-details}
\noindent\textbf{CF-aware methods.} Scratch (+Aug) \cite{garg2019counterfactual} minimizes the empirical cross-entropy loss computed using both original and counterfactual images.
CP \cite{russell2017worlds} has a regularization term that promotes the image pairs to be the same prediction. We use logits of a neural network model as representation vectors for the CP regularization term. Since Scratch (+aug) and CP utilize CTF samples, we generated these samples using IP2P with the same prompt in \cref{appendix:subsec_hyp_ip2p} using the image-CFG of 7.5 and the Text-CFG of 2.0. We note that we do not apply any filtering process for their generated training datasets. 
SenSeI \cite{yurochkin2020sensei} uses two metrics for training: one for a pre-defined fair regularizer distance metric and the other obtained by fair metric learning. We use the same metrics as presented in their code. By generating the worst-case samples based on these metrics, we apply a fair regularization term to promote their predictions to be the same, as originally implemented.
LASSI \cite{peychev2022latent} minimizes an objective function which is composed of the classification loss, the reconstruction loss, and the adversarial loss to learn individually fair representation. We use the official code of LASSI as it is.

\noindent\textbf{GF-aware methods.} LBC \cite{jiang2020identifying} necessitates multiple full-training iterations, alternately re-weighting each group based on the given group fairness metric and re-training. Due to its high computation budget for iterative full-training, we limit the number of epochs for each training to 5 and repeat this process 14 times. COV \cite{zafar2017fairness} utilizes a fairness constraint based on the covariance between the group label and the signed distance of feature vectors from the decision boundary of a classifier. We minimize the constraint-regularized objective function through gradient descent optimization, instead of directly solving its optimization problem. MFD \cite{jung2021mfd} employs an additional fairness-promoting regularization term based on Maximum Mean Discrepancy (MMD). For the MMD distance of the regularization term, we use the Gaussian RBF kernel with the variance parameter set as the mean of squared distance between all data points. We implemented SS \cite{idrissi2022simple} and RW \cite{kamiran2012data} identically to the original algorithm.

\noindent\textbf{CF- and GF-aware methods.}
The combinations of GF method and the augmentation were implemented so that GF methods train a model on their own objective function using training datasets augmented by generated CTF samples. The combinations of GF methods and CP optimize the objective functions of GF methods combined by the CP regularization. The CKD regularization term (\ref{eq:ckd}) builds upon representation vectors $f(\bm\theta, x)$. For this vector, we use the logits of a neural network model on LFW and CIFAR-10B. For CelebA, we utilize feature vectors from the penultimate layer of models as a representation vector since its training dataset is relatively much larger and more complex than others, leading to more fine-grained feature vectors. 

Our code is available at \url{https://github.com/sumin-yu/CKD.git}.

\subsection{Hyperparameter search}
The range of hyperparameter search used for \cref{table:method-comparison} and \cref{table:method-comparison-main} are shown in \cref{table:hyperparameters}. We utilize grid search to select hyperparameter values within a certain range. Note \oursmethod and the combinations of GF-aware methods and CP have additional parameters $\lambda$ for CP loss. We use the same range as CP for $\lambda$. We also note that for LASSI, we only search the hyperparameter for an adversarial loss while maintaining other parameters as the same as used in their experiments on CelebA.


\begin{table*}[th]
\centering
\caption{\small Hyperparameters and search ranges for each method.}
\vskip 0.15in
\label{table:hyperparameters}
\renewcommand{\arraystretch}{1.2}
\begin{tabular}{lcc}
\toprule
Method         & Hyperparameter & Search range \\
\midrule
CP {\scriptsize \cite{russell2017worlds}}    & CP strength $\lambda$ & $[10^{-2}, 10^2]$ \\
\midrule
SenSeI {\scriptsize \cite{yurochkin2020sensei}}   & Fair regularization strength $\rho$ & $[10^{-2}, 10^2]$ \\
\midrule
LASSI {\scriptsize \cite{peychev2022latent}}   & Adversarial loss weight $\lambda_2$ & $[10^{-3}, 10^{-1}]$ \\
\midrule
COV {\scriptsize \cite{zafar2017fairness}}  & Covariance strength $\lambda$ & $[10^{-2}, 10^2]$\\
\midrule
MFD {\scriptsize \cite{jung2021mfd}}            & MMD strength $\lambda$ & $[10^{-1}, 10^6]$\\
\midrule
LBC {\scriptsize \cite{jiang2020identifying}}           & LR for re-weights $\eta$ & $[10^{-1}, 10^3]$\\
\midrule
\oursmethod     & \oursmethod strength $\mu$ & $[10^{-1}, 10^3]$\\
\bottomrule
\end{tabular}
\vskip -0.1in
\end{table*}

\section{Rate of Flipped Predictions (RFP)}

\subsection{RFP measurement on CelebA}
\label{appendix:CD-G}

\begin{figure}[t!]
\vskip 0.1in
\begin{center}
\centerline{\includegraphics[width=0.95\columnwidth]{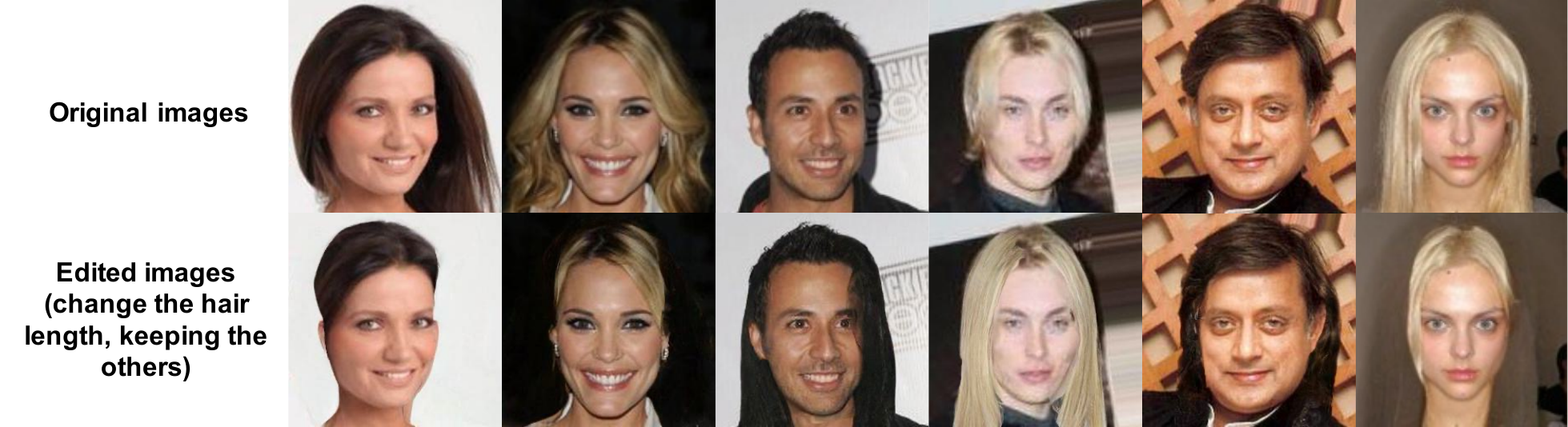}}
\caption{\small {\bf Examples of CelebA measuring RFP with respect to ``hair length''.} If the person in the original image (first row) had long hair, we create a modified image (second row) with shorter hair and vice versa.}
\label{fig:PC_G_example_CelebA}
\end{center}
\vskip -0.1in
\end{figure}

To measure RFP
on CelebA, we assume that the hair length of facial images is $G$, \ie, it is correlated with, but not caused by, the sensitive attribute. Then, 
we use Stochastic Differential Editing (SDEdit) \cite{meng2021sdedit}, an image editing method based on a diffusion model, to modify the hair length in each image. SDEdit selectively edits specific regions of a given image based on the colored stroke. Namely, SDEdit depicts the image region indicated by the stroke with the given color in the most plausible manner. By doing so, SDEdit generates realistic and faithful edited images, while preventing changes in the region not indicated by the strokes.
To utilize SDEdit, we randomly select 40 samples for each group of the same target label and sensitive attribute from original samples of CelebA-CTF pairs and then we manually apply strokes on the hair of facial images for a total of 160 samples. Specifically, To extend the hair length, we applied strokes with the hair color to the areas where the hair should grow. Conversely, to shorten the hair length, we applied strokes with the background color to the areas where the hair should be removed. After this process, we utilize the official PyTorch implementation of \citet{meng2021sdedit} to edit images with the applied strokes. \cref{fig:PC_G_example_CelebA} shows some examples of images edited by SDEdit.


\subsection{Discussion about selecting $G$ on CelebA}
\label{appendix:subsec_selection_g}
As mentioned in \cref{subsec:G_analysis}, we intuitively chose ``hair length'' as $G$ on CelebA since $G$ is highly correlated with, but not caused by, the sensitive attribute $A$.
However, we introduce a more generalized approach for choosing $G$ by leveraging a CP-trained model that exhibits low CD but high DEO. Specifically, using a pre-defined set of attributes, we first train a linear classifier on the top of the feature extractor from the CP-trained model for each attribute. In cases where annotations are not available, CLIP-based pseudo labels can be utilized. Based on the accuracy of each linear classifier, we can then identify which attributes the CP-trained model learns more, indicating potential heavy reliance on these attributes. Finally, we can select $G$ attributes based on two criteria: (1) high accuracy of a linear classifier and (2) high correlation (not causation) with the sensitive attribute.

To validate this approach, we conducted an experiment on CelebA dataset using a subset of 40 pre-annotated attributes and the ``hair length'' (''hair length'' labels are predicted by CLIP as it is not originally labeled in CelebA). \cref{table:G-selection} displays the accuracy for each attribute after training a linear classifier on the top of the CP-trained model, alongside the skewness between the sensitive attribute and other attributes. Attributes like ``Pale Skin'' show a high correlation with the sensitive attribute but low accuracies, suggesting CP might not rely on them (not satisfying the second condition). ``Mouth Sightly Open'' exhibits low correlation and accuracy, thus not being considered as $G$ (failing the first condition). In contrast, attributes such as ``bangs'', ``wearing hat'', and ``hair length'' exhibit both high correlation values and accuracies, indicating that they are promising candidates for the attribute $G$.
\begin{table*}[t!]
\centering
\small
\caption{The skewness between the sensitive attribute and other attributes, as well as the accuracy for each attribute after re-training a linear classifier on the top of the CP-trained model.}
\label{table:G-selection}
\vskip 0.15in
\begin{tabular}{lccccc}
\toprule
& Hair Length & Bangs & Wearing Hat & Pale Skin & Mouth Slightly Open \\
\midrule
\midrule
Acc (\%)     & 64.93 & 68.07 & 72.63 & 50.79 & 56.79 \\
\midrule
Skewness & 0.8 & 0.91  & 0.97 & 0.98 & 0.27 \\
\bottomrule
\end{tabular}
\vskip -0.1in
\end{table*}

\begin{figure*}[t!]
\vskip 0.1in
\begin{center}
\centerline{\includegraphics[width=1.0\columnwidth]{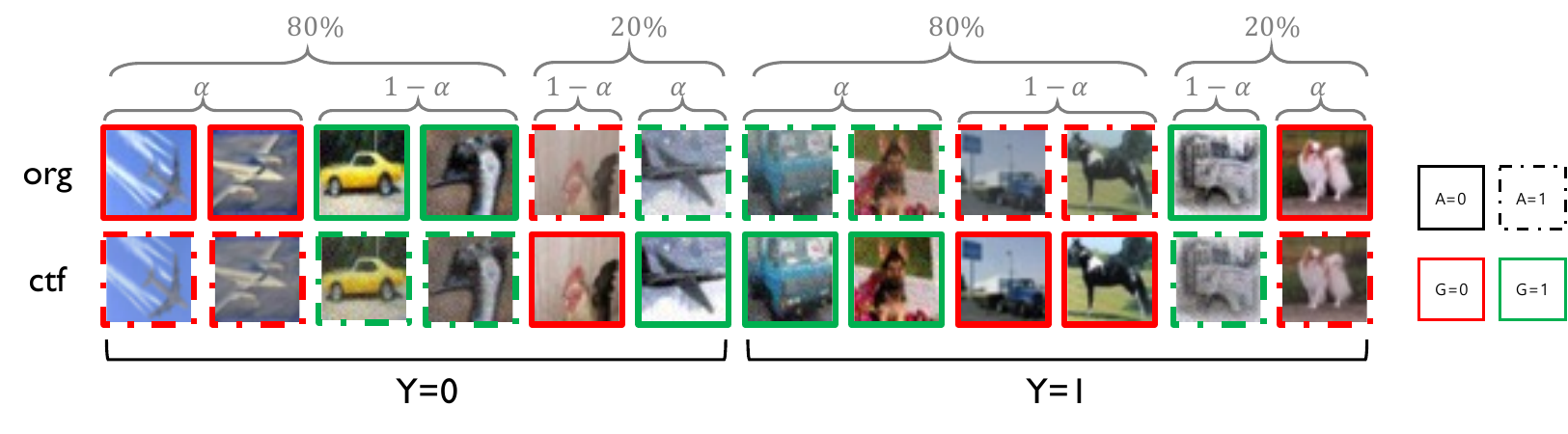}}
\caption{\small {\bf Illustration of CIFAR-10B.} The sensitive attribute $A$ is characterized by the line type ($A=0$ for a solid line, $A=1$ for a dashed line), while another attribute $G$ is denoted by the color of the line ($G=0$ for a red line, $G=1$ for a green line).}
\label{fig:CIFAR}
\end{center}
\vskip -0.1in
\end{figure*}

\section{Further discussion about the faithfulness assumption}
\label{appendix:faithful}
The faithful assumption states that if some two variables are statistically independent, they are d-separated, \ie, there is no connected path between them. Thus, under this assumption, GF-aware methods, which enforce independence between the sensitive attribute and the target label, can achieve CF simultaneously when they successfully achieve GF. The previous works \cite{anthis2023causal, rosenblatt2023counterfactual} provide the same argument that GF implies CF under the faithfulness assumption. However, some other previous works \cite{kennaway2020causation} have demonstrated that while the faithfulness assumption is crucial for causal inference literature, it may not always hold true, especially in complex real-world scenarios. Moreover, the result for GF-aware methods in \cref{table:method-comparison-main} reveals that the GF-aware methods have a minimal impact on improving CD, implying that the faithfulness assumption does not hold in CelebA and LFW datasets.

\section{Additional results}
\label{appendix:additional-results}

\begin{table}[t!]
    \centering
    \caption{\small \textbf{Evaluation of GF and CF of fair-training for image classification}. We put the results from     \cref{table:method-comparison} and \cref{table:method-comparison-main} together with the standard deviation values over four different seeds.}
    \label{table:method-comparison-std}
        \begin{subtable}[h]{0.6\textwidth}
        \small
        \centering
        \caption{\small Standard deviations on CIFAR-10B.}
        \setlength{\tabcolsep}{3.5pt}
        \renewcommand{\arraystretch}{0.8}
        \begin{tabular}{lccc}
        \toprule
        & \multicolumn{3}{c}{CIFAR-10B ($\alpha$=0.8)} \\
        Method  & Acc $\uparrow$  & CD $\downarrow$ & DEO $\downarrow$ \\        \midrule
        Scratch                                   & $78.01_{\pm 0.85}$ & $17.90_{\pm 2.43}$ & $27.46_{\pm 2.65}$ \\        \midrule
        \multicolumn{4}{c}{\small CF-aware training}\\ \midrule
        Scratch+aug \cite{garg2019counterfactual} & $75.38_{\pm 0.79}$ & $9.33_{\pm 0.88}$ & $15.25_{\pm 1.19}$ \\
        CP \cite{russell2017worlds}               & $75.26_{\pm 1.87}$ & $2.05_{\pm 0.15}$ & $33.23_{\pm 7.56}$ \\ 
        SenSeI \cite{yurochkin2020sensei}         & $77.21_{\pm 0.83}$ & $16.32_{\pm 1.23}$ & $24.18_{\pm 1.70}$ \\         \midrule
        \multicolumn{4}{c}{\small GF-aware training}\\ \midrule
        SS \cite{idrissi2022simple}    & $74.77_{\pm 0.41}$ & $16.42_{\pm 1.34}$ & $25.73_{\pm 2.12}$ \\
        RW \cite{kamiran2012data}      & $76.53_{\pm 0.57}$ & $12.15_{\pm 1.23}$ & $18.94_{\pm 1.71}$ \\
        COV \cite{zafar2017fairness}   & $79.03_{\pm 0.49}$ & $13.90_{\pm 0.39}$ & $24.05_{\pm 1.09}$ \\
        MFD \cite{jung2021mfd}         & $76.84_{\pm 0.92}$ & $12.24_{\pm 1.51}$ & $15.39_{\pm 1.28}$ \\
        LBC \cite{jiang2020identifying}& $76.16_{\pm 1.36}$ & $15.01_{\pm 2.26}$ & $17.12_{\pm 4.68}$ \\        \midrule
        \multicolumn{4}{c}{\small both CF and GF-aware training}\\ \midrule
        SS+aug  & $73.45_{\pm 0.46}$ & $9.95_{\pm 0.58}$ & $15.21_{\pm 2.06}$ \\
        RW+aug  & $76.15_{\pm 0.52}$ & $12.93_{\pm 0.96}$ & $20.94_{\pm 2.57}$ \\
        COV+aug & $76.52_{\pm 0.56}$ & $8.17_{\pm 0.22}$ & $15.04_{\pm 0.96}$ \\
        MFD+aug & $77.10_{\pm 0.58}$ & $11.16_{\pm 1.08}$ & $14.79_{\pm 2.15}$ \\
        LBC+aug & $75.82_{\pm 0.54}$ & $9.01_{\pm 0.67}$ & $15.29_{\pm 0.56}$ \\       \midrule
        SS+CP                       & $76.54_{\pm 1.48}$ & $3.14_{\pm 0.26}$ & $9.08_{\pm 1.22}$ \\
        RW+CP                       & $75.68_{\pm 0.37}$ & $8.83_{\pm 0.66}$ & $13.92_{\pm 0.94}$ \\
        COV+CP                      & $77.74_{\pm 0.52}$ & $4.30_{\pm 0.33}$ & $19.42_{\pm 1.74}$ \\
        MFD+CP                     & $76.67_{\pm 1.00}$ & $10.01_{\pm 0.57}$ & $13.17_{\pm 0.95}$ \\
        LBC+CP                      & $76.88_{\pm 2.64}$ & $3.02_{\pm 1.29}$ & $12.45_{\pm 1.93}$ \\        \midrule        \midrule
        \oursmethod($\lambda=0$)  & $76.32_{\pm 0.59}$ & $8.59_{\pm 1.32}$ & $11.23_{\pm 1.04}$ \\
        \oursmethod               & $78.49_{\pm 0.66}$ & $2.85_{\pm 0.20}$ & $7.30_{\pm 0.46}$ \\
        \bottomrule
        \end{tabular}
        \end{subtable}
        \begin{subtable}[h]{0.95\textwidth}
        \small
        \centering
        \caption{\small Standard deviations on CelebA and LFW.}
        \setlength{\tabcolsep}{3.5pt}
        \renewcommand{\arraystretch}{0.8}
        \begin{tabular}{lccccccc}
        \toprule
        & \multicolumn{3}{c}{CelebA (and \oursceleb)} && \multicolumn{3}{c}{LFW (and \ourslfw)}  \\
        Method         & Acc $\uparrow$  & CD $\downarrow$ & DEO $\downarrow$ && Acc $\uparrow$  & CD $\downarrow$ & DEO $\downarrow$ \\        \midrule
        Scratch                                   & $95.53_{\pm 0.06}$   & $10.26_{\pm 1.33}$  & $47.10_{\pm 5.57}$  && $90.85_{\pm 0.27}$  & $18.06_{\pm 1.89}$ & $7.66_{\pm 0.49}$  \\
        \midrule
        \multicolumn{8}{c}{\small CF-aware training}\\ \midrule
        Scratch+aug \cite{garg2019counterfactual} & $95.41_{\pm 0.15}$  & $4.65_{\pm 0.86}$  & $44.71_{\pm 2.57}$ && $90.34_{\pm 0.58}$   & $12.15_{\pm 1.29}$  & $7.86_{\pm 1.70}$  \\
        CP \cite{russell2017worlds}               & $94.10_{\pm 0.08}$ & $2.53_{\pm 1.26}$ & $51.01_{\pm 1.71}$ && $89.77_{\pm 0.90}$ & $9.20_{\pm 0.56}$ & $8.74_{\pm 1.43}$ \\
        SenSeI \cite{yurochkin2020sensei}         & $95.33_{\pm 0.30}$  & $8.00_{\pm 0.59}$  & $52.32_{\pm 5.26}$ && $87.75_{\pm 3.78}$   & $16.09_{\pm 3.70}$   & $9.23_{\pm 1.38}$ \\ 
        LASSI \cite{peychev2022latent}            & $91.07_{\pm 0.27}$ &  $9.69_{\pm 0.78}$ & $31.79_{\pm 3.17}$ && - & - & - \\        \midrule
        \multicolumn{8}{c}{\small GF-aware training}\\ \midrule
        SS \cite{idrissi2022simple}    & $95.44_{\pm 0.09}$ &  $9.13_{\pm 2.73}$ & $42.95_{\pm 3.87}$ && $90.43_{\pm 0.21}$ & $18.19_{\pm 1.27}$ & $6.75_{\pm 0.33}$ \\
        RW \cite{kamiran2012data}      & $95.16_{\pm 0.08}$ & $5.50_{\pm 0.46}$ & $24.21_{\pm 1.76}$ && $90.87_{\pm 0.25}$ & $18.68_{\pm 2.91}$ & $6.92_{\pm 0.96}$ \\
        COV \cite{zafar2017fairness}   & $94.42_{\pm 0.16}$ & $7.72_{\pm 2.89}$ & $34.04_{\pm 4.43}$ && $90.85_{\pm 0.50}$ & $16.43_{\pm 2.12}$ & $6.99_{\pm 1.23}$ \\
        MFD \cite{jung2021mfd}         & $94.37_{\pm 0.77}$ & $4.61_{\pm 1.77}$ & $19.00_{\pm 6.44}$ && $90.47_{\pm 0.10}$ & $16.07_{\pm 2.01}$ & $2.15_{\pm 0.51}$ \\
        LBC \cite{jiang2020identifying}& $94.92_{\pm 0.28}$ & $6.24_{\pm 0.69}$ & $22.61_{\pm 1.79}$ && $90.71_{\pm 0.61}$ & $15.76_{\pm 0.86}$ & $3.56_{\pm 2.02}$  \\        \midrule
        \multicolumn{8}{c}{\small both CF and GF-aware training}\\ \midrule
        SS+aug  & $95.17_{\pm 0.02}$ & $5.24_{\pm 1.02}$ & $40.80_{\pm 2.86}$ && $89.96_{\pm 0.24}$ & $15.23_{\pm 2.21}$ & $6.82_{\pm 0.88}$ \\
        RW+aug  & $95.13_{\pm 0.04}$ & $5.34_{\pm 0.59}$ & $24.63_{\pm 1.58}$ && $90.76_{\pm 0.16}$ & $18.63_{\pm 2.07}$ & $6.71_{\pm 1.38}$ \\
        COV+aug & $94.08_{\pm 0.41}$ & $8.11_{\pm 2.26}$ & $29.03_{\pm 0.72}$ && $90.47_{\pm 0.20}$ & $13.65_{\pm 1.71}$ & $6.78_{\pm 0.09}$ \\
        MFD+aug & $93.78_{\pm 0.80}$ & $3.87_{\pm 0.84}$ & $14.36_{\pm 4.39}$ && $89.90_{\pm 0.62}$ & $19.36_{\pm 2.41}$ & $2.47_{\pm 0.75}$ \\
        LBC+aug & $94.39_{\pm 1.44}$ & $9.32_{\pm 4.20}$ & $36.08_{\pm 11.36}$ && $88.66_{\pm 1.25}$ & $12.41_{\pm 2.02}$ & $2.79_{\pm 1.36}$\\ \midrule
        SS+CP      & $94.54_{\pm 0.09}$ & $2.40_{\pm 0.35}$ & $37.97_{\pm 2.27}$ && $88.70_{\pm 0.82}$ & $6.13_{\pm 1.17}$ & $4.26_{\pm 1.74}$ \\
        RW+CP      & $95.19_{\pm 0.13}$ & $4.67_{\pm 0.76}$ & $25.56_{\pm 2.87}$ && $90.87_{\pm 0.28}$ & $15.24_{\pm 1.67}$ & $6.16_{\pm 0.19}$ \\
        COV+CP     & $94.29_{\pm 0.18}$ & $5.36_{\pm 1.00}$ & $51.63_{\pm 0.67}$ && $91.23_{\pm 0.37}$ & $11.91_{\pm 2.18}$ & $6.52_{\pm 1.05}$  \\
        MFD+CP     & $93.81_{\pm 0.30}$ & $3.47_{\pm 0.52}$   & $23.31_{\pm 0.74}$ && $89.39_{\pm 1.90}$   & $15.15_{\pm 1.24}$ & $1.90_{\pm 1.03}$  \\
        LBC+CP     & $95.12_{\pm 0.10}$   & $4.72_{\pm 0.87}$   & $22.78_{\pm 2.26}$ && $89.92_{\pm 0.28}$  & $8.33_{\pm 1.07}$ & $3.02_{\pm 0.58}$ \\        \midrule \midrule
        \oursmethod($\lambda=0$)& $94.12_{\pm 0.23}$ & $4.31_{\pm 1.47}$ & $14.11_{\pm 1.25}$ && $90.76_{\pm 0.13}$ & $12.42_{\pm 2.69}$ & $2.64_{\pm 0.14}$ \\
        \oursmethod             & $93.08_{\pm 0.46}$  & $4.44_{\pm 0.70}$ & $13.23_{\pm 1.30}$ && $89.26_{\pm 0.45}$ & $7.94_{\pm 0.89}$ & $1.88_{\pm 0.67}$ \\
        \bottomrule
        \end{tabular}
        \end{subtable}
        \end{table}
\begin{table}[t]
\centering
\small
\caption{\small {\bf Impact of robustness to $G$ of the teacher model on CKD with Celeb and LFW.}
$\bm\theta^T_\text{CKD}$ and $\bm\theta^T_\text{CP}$ are CKD and CP trained teacher models. $\bm\theta^T_\text{Scratch}$ is a vanilla-trained teacher model. RFP$^T$ denotes how a teacher is biased towards $G$. DEO, CD, RFP are metrics for evaluating GF, CF, and bias towards $G$, respectively. Since we generate a dataset for RFP measurements on CelebA with ``hair length'' as $G$, we report RFP values only for CelebA.}
\vskip 0.1in
\label{table:change-teacher-real}
\setlength{\tabcolsep}{4pt}
\begin{tabular}{lccccccccc}
\toprule
 & \multicolumn{5}{c}{CelebA} && \multicolumn{3}{c}{LFW} \\
Method & RFP$^T$ $\downarrow$ & Acc $\uparrow$  & DEO $\downarrow$ & CD $\downarrow$ & RFP $\downarrow$ && Acc $\uparrow$  & DEO $\downarrow$ & CD $\downarrow$  \\
\midrule
\midrule
\oursmethod w/ $\bm\theta^T_\text{Scratch}$ & 15.27 & 93.08 & 13.23 & 4.44 & 10.85  && 89.26 & 1.88 & 7.94 \\
\midrule
\oursmethod w/ $\bm\theta^T_\text{CKD}$ & 10.85 & 93.98 & 14.37 & 4.05 & 11.64 && 89.17 & 1.48 & 8.07  \\
\oursmethod w/ $\bm\theta^T_\text{CP}$  & 20.37 & 94.25 & 34.49 & 3.28 & 16.61 && 89.85 & 9.26 & 8.32   \\
\bottomrule
\end{tabular}
\vskip -0.1in
\end{table}

\subsection{Result tables with standard deviation}
\label{appendix:subsec_std}
We report the standard deviation values of the performance comparison results in \cref{table:method-comparison-std} for CIFAR-10B, CelebA, and LFW respectively. The standard deviation values are calculated over four different seeds.

\subsection{Impact of the robustness to $G$ of the teacher model on \oursmethod with CelebA and LFW}
\label{appdendix:robustness-teacher}
We analyze the effectiveness of the teacher model on real image datasets. We report the performance of CKD using variants of teacher models that are more or less robust to $G$ on CelebA and LFW. We consider three teacher models, ordered by robustness to $G$: CP ($\bm\theta^T_\text{CP}$), Scratch ($\bm\theta^T_\text{Scratch}$), and CKD model with a Scratch teacher ($\bm\theta^T_\text{CKD}$). \cref{table:change-teacher-real} displays ACC, DEO, CD (on CelebA and LFW), and RFP (on CelebA) depending on the teacher models. Since we generate a dataset for RFP measurements on CelebA with ``hair length'' as $G$, we report RFP values only for CelebA. Through the result, we observe that compared to vanilla-trained teachers $\bm\theta^T_\text{Scratch}$, using more robust teachers (\eg, $\bm\theta^T_\text{CKD}$) achieves slightly better or competitive DEO and CD, while employing less robust teachers (\eg, $\bm\theta^T_\text{CP}$) significantly degrades DEO, which are consistent with the results in \cref{subsec:teacher_analysis} on CIFAR-10B.

\subsection{Analaysis on CKD}
\label{subsec:appendix_ckd_abl}
\noindent\textbf{Ablation study.} 
To study the effectiveness of our CKD regularization term, we additionally consider a method that is a naive combination of a typical KD method proposed by \citet{hinton2015distilling} and CP \cite{russell2017worlds} (\ie, HKD+CP). Note that this combination can be considered as a baseline method that considers both CF and GF if it uses a robust teacher model to $G$, because the method robustifies the model with respect to $G$ while achieving CF. \cref{table:method-comparison-appendix} compares the method with our CKD. As we expected, the results show that HKD+CP improves both DEO and CD simultaneously. However, its performance is still suboptimal compared to CKD, showing CKD is more effective than the naive combination of KD and CP.


\noindent\textbf{CKD on feature vectors vs logits.}
CKD can utilize either feature or logit vectors as the target vectors, i.e., $f(\theta, x)$. For CIFAR-10B and LFW, where we use logits as the target vectors in our experiments, we displayed the performance of CKD using feature vectors as the target vectors in \cref{table:CKD-vectors}. The results demonstrate that CKD using feature vectors exhibits comparable performance to those with logits. Moreover, we can get even better performance on CIFAR-10B using feature vectors, demonstrating that CF and GF can be achieved simultaneously regardless of which type of target vectors is used.

\noindent\textbf{Sensitiveness of $\mu$.}
$\mu$ is the regularization strength for the CKD loss term. Specifically, as $\mu$ increases, we expect improvements in both DEO and CD. \cref{table:sensitivity-mu} shows the performance of CKD across different values of $\mu$, aligning with our expectations. Additionally, we note that CKD performance is insensitive to $\mu$. 

\noindent\textbf{The implication of using non-curated IP2P.}
Uncurated CTF datasets are imperfect. Specifically, some samples generated from the original images in our test datasets were filtered out because the images either showed minimal changes or had alterations that affected non-sensitive attributes including the target attributes. Consequently, the more such incomplete samples exist, the more they will negatively impact the performance of our method.
To assess how sensitive CKD is to incomplete CTF training samples, we conducted additional experiments on CIFAR-10B by varying the ratio of incomplete CTF samples in the training set. For a given ratio $\alpha$ , we assumed that half of the incomplete samples are nearly unchanged, while the other half are samples where both the target and sensitive attributes are altered. We varied $\alpha$ from 20\% to 60\% in 10\% increments and reported the accuracy, CD, and DEO of CKD in the table below. The results in \cref{table:non-curated-IP2P} indicate that CKD significantly improves both CD and DEO compared to Scratch, even for high $\alpha$ s. Although this phenomenon has not been fully explained, we hypothesize that the robustness can be attributed to the distillation process, as empirically demonstrated in \cite{goldblum2020adversarially}.

\subsection{Additional experimental results on counterfactual samples.}
We additionally report the accuracy (acc-CTF) and DEO (DEO-CTF) for CKD and several baseline methods on counterfactual samples in CelebA-CF in \cref{table:acc-deo-on-CTF}.

\subsection{Additional experimental results for other metrics proposed by \cite{pinto2024matrix}.}
\label{appendix:additional-metric-CCM}
We first note that the Counterfactual Disparity (CD) we used is the same metric as the Switch Rate (SR) proposed by \cite{pinto2024matrix}. We computed P2NR (another metric proposed by \cite{pinto2024matrix}) on CelebA and obtained values of $0.036$, $0.165$, and $0.339$ for Scratch, CP, and CKD, respectively. These results indicate that CKD achieves low CD with a balanced rate of misclassification across the labels.
Additionally, we would like to emphasize that \citeauthor{pinto2024matrix} focused on scenarios where GF does not imply CF in their experiments—highlighting cases where the faithfulness assumption, which can be overly stringent, does not hold (see line 323 in their paper). However, our work primarily explores the converse: whether CF can imply GF depending on the presence of $G$, independent of the faithfulness assumption.

\begin{table}[t!]
\centering
\small
\caption{\small \textbf{The implication of using non-curated IP2P.}}
\label{table:non-curated-IP2P}
\vskip 0.15in
\begin{tabular}{cccc}
\toprule
  & Acc $\uparrow$  & CD $\downarrow$ & DEO $\downarrow$  \\
\midrule
\midrule
Scratch & 78.01 & 17.90 & 27.46 \\
CKD & 78.49 & 2.85 & 7.30 \\
CKD (20\%) & 79.82 & 2.77 & 11.41 \\
CKD (30\%) & 79.76 & 2.88 & 12.78 \\
CKD (40\%) & 79.70 & 2.94 & 12.88 \\
CKD (50\%) & 79.72 & 2.85 & 14.10 \\
CKD (60\%) & 79.61 & 3.01 & 14.94 \\
\bottomrule
\end{tabular}
\vskip -0.1in
\end{table}

\begin{table}[t!]
\centering
\small
\caption{\small \textbf{The accuracy and DEO on counterfactual samples in CelebA-CF.}}
\label{table:acc-deo-on-CTF}
\vskip 0.15in
\begin{tabular}{ccc}
\toprule
  & Acc-CTF $\uparrow$  & DEO-CTF $\downarrow$  \\
\midrule
\midrule
Scratch & 77.22 & 15.92 \\
SS & 81.43 & 28.74 \\
RW & 79.75 & 23.61 \\
LBC & 78.06 & 33.95 \\
CP & 75.21 & 62.82 \\
CKD & 90.08 & 21.86 \\
\bottomrule
\end{tabular}
\vskip -0.1in
\end{table}

\begin{table}[t!]
\centering
\small
\caption{\small \textbf{Evaluation of group fairness (GF) and counterfactual fairness (CF) of fair-training for image classification}. The details are the same as \cref{table:method-comparison-main}. ``HKD+CP'' denotes a model that naively combines Knowledge Distillation \cite{hinton2015distilling} with CP \cite{russell2017worlds}.}
\setlength{\tabcolsep}{4.5pt}
\label{table:method-comparison-appendix}
\begin{tabular}{lccccccccccc}
\toprule
& \multicolumn{3}{c}{CIFAR-10B ($\alpha$=0.8)} && \multicolumn{3}{c}{CelebA (and \oursceleb)} && \multicolumn{3}{c}{LFW (and \ourslfw)} \\
Method         & Acc $\uparrow$ & DEO $\downarrow$ & CD $\downarrow$ && Acc $\uparrow$ & DEO $\downarrow$ & CD $\downarrow$ && Acc $\uparrow$ & DEO $\downarrow$ & CD $\downarrow$ \\
\midrule
Scratch        & 78.36 & 27.28 & 17.68 && 95.53   & 47.10    & 10.36  && 90.85  & 7.66   & 18.06 \\
\midrule
HKD+CP  & 79.18 & 16.54 & 2.40 && 93.95 & 33.98 & 4.40  && 89.11 & 3.76 & 8.78   \\
\midrule
\midrule
\oursmethod($\lambda=0$)  & 75.63 & 6.33 & 8.94 && 94.12 & 14.11 & 4.31 && 90.76 & 2.64 & 12.47 \\
\oursmethod    & 78.46 & 7.11 & 2.86 && 93.08 & 13.23  & 4.44   && 89.26  & 1.88   & 7.94  \\
\bottomrule
\end{tabular}
\end{table}

\begin{table}[t!]
\centering
\small
\caption{\small \textbf{CKD on feature vectors vs logits.}}
\label{table:CKD-vectors}
\vskip 0.15in
\begin{tabular}{lcccccc}
\toprule
& \multicolumn{3}{c}{CIFAR-10B ($\alpha$=0.8)} & \multicolumn{3}{c}{LFW} \\
Method  & Acc $\uparrow$  & DEO $\downarrow$ & CD $\downarrow$  & Acc $\uparrow$  & DEO $\downarrow$ & CD $\downarrow$  \\
\midrule
\midrule
\oursmethod w/ logit                   & 78.46 & 7.11 & 2.86 & 89.26 & 1.88 & 7.94 \\
\midrule
\midrule
\oursmethod w/ feature  &   78.24   & 2.64 & 1.23 & 88.37 & 3.98 & 6.22  \\
\bottomrule
\end{tabular}
\vskip -0.1in
\end{table}

\begin{table}[t!]
\centering
\small
\caption{\small \textbf{Sensitivity of $\mu$.}}
\label{table:sensitivity-mu}
\vskip 0.15in
\begin{tabular}{ccccccccc}
\toprule
& \multicolumn{4}{c}{CelebA} & \multicolumn{4}{c}{LFW} \\
$\mu$  & Acc $\uparrow$  & DEO $\downarrow$ & CD $\downarrow$  && $\mu$ & Acc $\uparrow$  & DEO $\downarrow$ & CD $\downarrow$  \\
\midrule
\midrule
 0.01  &  94.13  &  14.83 &  4.4 &      &  0.01  &  90.05 &  2.82 &  15.01 \\
 0.1  & 94.13 & 14.45 & 3.03 &     &  0.1  & 90.09 & 2.39 & 14.19 \\
 1.0  & 93.63 & 13.84 & 4.24  &     &  1.0  & 89.78 & 2.75 & 13.66 \\
 7.0  & 93.08 & 13.23 & 4.44 &     & 10.0  & 89.23 & 1.85 & 13.82 \\
 10.0 & 92.93 & 13.67 & 4.10  &     & 50.0  & 88.83 & 1.8  & 7.80  \\
\bottomrule
\end{tabular}
\vskip -0.1in
\end{table}

\clearpage
\section{Datasheet for dataset}
\label{appendix:datasheet}

\subsection{Motivation}
 \textbf{For what purpose was the dataset created?} These datasets were created for evaluating counterfactual fairness in image classifiers. Furthermore, since our datasets contain counterfactual images generated from real-world images, our datasets can be also used for analyzing the relationship between counterfactual and group fairness on image datasets. For more discussion of the motivation behind our datasets, see \cref{sec:intro}.

 \textbf{Who created the dataset (e.g., which team, research group) and on behalf of which entity (e.g., company, institution, organization)?}  The datasets were created by the authors of this paper who were affiliated with Seoul National University and NAVER AI LAB. 

 \textbf{Who funded the creation of the dataset?} Funding was provided by the National Research Foundation of Korea (NRF); Institute of Information \& Communications Technology Planning \& Evaluation (IITP); and the SNU-Naver Hyperscale AI Center.

\subsection{Composition}
 \textbf{What do the instances that comprise the dataset
    represent (e.g., documents, photos, people, countries)?} The instances represent synthetically generated images and corresponding real-world original images from two popular benchmark facial image datasets, CelebA \cite{celeba} and LFW \cite{Huang2007a}. 

 \textbf{How many instances are there in total (of each type, if appropriate)?}
 CelebA-CF and LFW-CF contain a total of 230 and 144 image pairs of original and counterfactual images, respectively.

 \textbf{Does the dataset contain all possible instances or is it
    a sample (not necessarily random) of instances from a larger set?}
  We uniformly sampled a subset of test images in CelebA and LFW to balance the target and group labels (see more details in \cref{sec:ctf_contruction}). Then, we made our datasets including all possible samples according to our filtering process.

 \textbf{What data does each instance consist of?} Each instance contains a pair of original and counterfactual images. 

 \textbf{Is there a label or target associated with each
    instance?} Yes, there are 40 binary annotations that originated from CelebA and LFW.

 \textbf{Is any information missing from individual instances?} No

 \textbf{Are relationships between individual instances made
    explicit (e.g., users' movie ratings, social network links)?} Yes, instances that correspond to a counterfactual pair are explicitly annotated as such in our dataset. Otherwise, there are no relationships between individual instances.

 \textbf{Are there recommended data splits (e.g., training,
    development/validation, testing)?} No, the dataset is created for the purpose of testing.

 \textbf{Are there any errors, sources of noise, or redundancies
    in the dataset?} Using IP2P as the initial step in constructing our datasets might introduce some noise or errors in the datasets. Refer to \cref{appendix:limitation} for further details.

 \textbf{Is the dataset self-contained, or does it link to or
    otherwise rely on external resources (e.g., websites, tweets,
    other datasets)?} Yes, it is self-contained.

 \textbf{Does the dataset contain data that might be considered
    confidential (e.g., data that is protected by legal privilege or
    by doctor patient confidentiality, data that includes the content of individuals' non-public communications)?} No

 \textbf{Does the dataset contain data that, if viewed directly,
    might be offensive, insulting, threatening, or might otherwise
    cause anxiety?} Yes, the dataset may cause some anxiety about sex labels. See \cref{sec:ctf_contruction} and \cref{appendix:limitation}.

 \textbf{Does the dataset identify any subpopulations (e.g., by
    age, gender)?} Yes, our datasets were created after evaluating whether counterfactual samples regarding visually perceived sexual traits were generated correctly or not. This evaluation was conducted by five human annotators. Thus, our datasets contain the identification of visually perceived sexual traits which represent some statistically representative features for each sex. See more discussion in \cref{sec:ctf_contruction}.

 \textbf{Is it possible to identify individuals (i.e., one or
    more natural persons), either directly or indirectly (i.e., in
    combination with other data) from the dataset?} Yes, our datasets are generated from CelebA and LFW, which are facial datasets collected on the internet. 

 \textbf{Does the dataset contain data that might be considered
    sensitive in any way (e.g., data that reveals race or ethnic
    origins, sexual orientations, religious beliefs, political
    opinions or union memberships, or locations; financial or health
    data; biometric or genetic data; forms of government
    identification, such as social security numbers; criminal history)?} 
    Yes, we set sex as the sensitive attribute and created our CTF samples with the sensitive attribute flipped. 

\subsection{Collection Process}

 \textbf{How was the data associated with each instance
    acquired?} Our datasets are generated through image editing using IP2P (See \cref{sec:ctf_contruction}).

 \textbf{What mechanisms or procedures were used to collect the
    data (e.g., hardware apparatuses or sensors, manual human
    curation, software programs, software APIs)?} Refer to \cref{sec:ctf_contruction} for a complete description of our data generation process.

 \textbf{If the dataset is a sample from a larger set, what was
    the sampling strategy (e.g., deterministic, probabilistic with
    specific sampling probabilities)?}
    Original test samples in our datasets were uniformly sampled from the test datasets of CelebA and LFW.

 \textbf{Who was involved in the data collection process (e.g.,
    students, crowdworkers, contractors) and how were they compensated
    (e.g., how much were crowdworkers paid)?} The filtering process for our datasets involved five student annotators who received about 18 USD per hour for their wage.

 \textbf{Over what timeframe was the data collected?} Our datasets were generated and evaluated over one month.

 \textbf{Were any ethical review processes conducted (e.g., by an
    institutional review board)?} No

 \textbf{Did you collect the data from the individuals in
    question directly, or obtain it via third parties or other sources
    (e.g., websites)?} No, we initially obtained the data from publicly available sources. Subsequently, we edited the data and filtered the edited one through human annotators.

 \textbf{Were the individuals in question notified about the data
    collection?} Not applicable

 \textbf{Did the individuals in question consent to the
    collection and use of their data?} Not applicable

 \textbf{If consent was obtained, were the consenting individuals
    provided with a mechanism to revoke their consent in the future or
    for certain uses?} Not applicable

 \textbf{Has an analysis of the potential impact of the dataset and its use on data subjects (e.g., a data protection impact analysis) been conducted?} Not applicable

\subsection{Preprocessing/cleaning/labeling}

 \textbf{Was any preprocessing/cleaning/labeling of the data done
    (e.g., discretization or bucketing, tokenization, part-of-speech
    tagging, SIFT feature extraction, removal of instances, processing
    of missing values)?} Yes, we filtered our generated datasets with human annotators. Refer to \cref{sec:ctf_contruction} for a complete description of our filtering process.

 \textbf{Was the ``raw'' data saved in addition to the preprocessed/cleaned/labeled data (e.g., to support unanticipated future uses)?} No, however, raw data can be reproduced by applying IP2P as described in \cref{sec:ctf_contruction}.

 \textbf{Is the software that was used to preprocess/clean/label the data available?} Yes, refer to the \cref{sec:ctf_contruction}.

\subsection{Uses}

 \textbf{Has the dataset been used for any tasks already?} Yes, we applied our datasets to evaluate CF in image classifiers in \cref{sec:pre-exp} and \ref{sec:exp} and analyze the relationship between CF and GF in \cref{sec:G_analysis}.

 \textbf{Is there a repository that links to any or all papers or systems that use the dataset?}
 We will provide a link to a repository on GitHub that includes references to all papers and systems utilizing the dataset.

 \textbf{What (other) tasks could the dataset be used for?} There is no other task where our dataset can be used. The dataset is exclusively designed for evaluating counterfactual fairness in real-world image datasets.

 \textbf{Is there anything about the composition of the dataset or the way it was collected and preprocessed/cleaned/labeled that might impact future uses?} Because our datasets were generated through the image editing technique, IP2P \cite{brooks2023instructpix2pix}, they may contain implicit biases or errors, which are present in the IP2P model \cite{luccioni2023stable, Bianchi_2023}. While we have conducted a thorough human filtering and validation process to minimize these issues in our dataset, future users should still be aware of these limitations.

 \textbf{Are there tasks for which the dataset should not be used?} The dataset should not be employed for tasks where the limitations discussed in \cref{appendix:limitation} could pose critical issues, or for tasks that are not for research purposes.

\subsection{Distribution}

 \textbf{Will the dataset be distributed to third parties outside of the entity (e.g., company, institution, organization) on behalf of which the dataset was created?} Yes, the datasets will be made publicly available.

 \textbf{How will the dataset will be distributed (e.g., tarball on website, API, GitHub)?} The dataset will be distributed using tarball on the website. Refer to \cref{appendix:our-dataset}.

 \textbf{When will the dataset be distributed?} The datasets will be made publicly available upon acceptance.

 \textbf{Will the dataset be distributed under a copyright or other intellectual property (IP) license, and/or under applicable terms of use (ToU)?} The datasets will be distributed under the CC BY 4.0 license.

 \textbf{Have any third parties imposed IP-based or other restrictions on the data associated with the instances?} 
No
 
 \textbf{Do any export controls or other regulatory restrictions apply to the dataset or to individual instances?} No

\subsection{Maintenance}
 
 \textbf{Who will be supporting/hosting/maintaining the dataset?} The datasets are hosted, supported, and maintained by the authors.

 \textbf{How can the owner/curator/manager of the dataset be contacted (e.g., email address)?} The corresponding author can be contacted by the e-mail address which will be listed on the first page of this paper after camera-ready.

 \textbf{Is there an erratum?} No

 \textbf{Will the dataset be updated (e.g., to correct labeling
    errors, add new instances, delete instances)?} No future updates are currently planned. However, we will monitor the GitHub repository for related issues and address any problems that arise.

 \textbf{If the dataset relates to people, are there applicable
    limits on the retention of the data associated with the instances
    (e.g., were the individuals in question told that their data would be retained for a fixed period of time and then deleted)?} Not applicable

 \textbf{Will older versions of the dataset continue to be
    supported/hosted/maintained?} Yes, if the datasets are updated, we will maintain the older versions.

 \textbf{If others want to extend/augment/build on/contribute to
    the dataset, is there a mechanism for them to do so?} Yes, we make our code and datasets public, and hence others can contribute or extend to our work and datasets.

\clearpage

\begin{table}[t]
\small
\centering
\caption{\small \textbf{Evaluation of GF and CF of fair-training for image classification}.}
\setlength{\tabcolsep}{8pt}
\label{table:method-comparison-main}
\begin{tabular}{lcccccc}
\toprule
& \multicolumn{3}{c}{CelebA (and \oursceleb)} & \multicolumn{3}{c}{LFW (and \ourslfw)} \\
Method         & Acc $\uparrow$ & CD $\downarrow$ & DEO $\downarrow$ & Acc $\uparrow$  & CD $\downarrow$ & DEO $\downarrow$\\
\midrule
Scratch        & 95.53   & 10.26  & 47.10    & 90.85  & 18.06 & 7.66   \\
\midrule
Scratch+aug  & 95.41    & 4.65  & {44.71}   & 90.34   & 12.15  & {7.86} \\
CP  & 94.10   & \textbf{2.53} & {51.01}    & 89.77  & 9.20 & {8.74}  \\ \midrule
RW  & 95.16 & 5.50 & 24.21 & 90.87 & {18.68} & 6.92 \\
COV & 94.42 & 7.72 & 34.04 & 90.85 & 16.43 & 6.99 \\
LBC & 94.92   & 6.24   & \textbf{22.61}   & 90.71  & 15.76 & 3.56   \\
\midrule
RW+aug & 95.13 & 5.34 & 24.63 & 90.76 & {18.63} & 6.71 \\
COV+aug & 94.08 & 8.11 & 29.03 & 90.47 & 13.65 & 6.78 \\
LBC+aug & 94.39 & 9.32 & 36.08 & 88.66 & 12.41 & \textbf{2.79} \\
RW+CP & 95.19 & 4.67 & 25.56 & 90.87 & 15.24 & 6.16 \\
COV+CP & 94.29 & 5.36 & {51.63} & 91.23 & 11.91 & 6.52 \\
LBC+CP      & 95.12   & 4.72   & 22.78   & 89.92  & \textbf{8.33} & 3.02   \\
\midrule
\midrule
\oursmethod    & 93.08  & \textbf{4.44}    & \textbf{13.23}   & 89.26  & \textbf{7.94}  & \textbf{1.88}   \\
\bottomrule
\end{tabular}
\end{table}

\end{document}